%% file: paper.tex
\documentclass{article}

\PassOptionsToPackage{usenames,dvipsnames}{xcolor}

\usepackage[accepted]{icml2023}
\input{preamble/preamble}

\usepackage[textsize=tiny]{todonotes}

\icmltitlerunning{Regularizing towards soft equivariance under mixed symmetries}

\begin{document}

\twocolumn[
\icmltitle{Regularizing Towards Soft Equivariance Under Mixed Symmetries}



\icmlsetsymbol{equal}{*}

\begin{icmlauthorlist}
\icmlauthor{Hyunsu Kim}{gsai}
\icmlauthor{Hyungi Lee}{gsai}
\icmlauthor{Hongseok Yang}{soc,gsai,dimag}
\icmlauthor{Juho Lee}{gsai,aitrics}
\end{icmlauthorlist}

\icmlaffiliation{gsai}{Kim Jaechul Graduate School of AI, KAIST, Daejeon, South Korea}
\icmlaffiliation{soc}{School of Computing, KAIST, Daejeon, South Korea}
\icmlaffiliation{dimag}{Discrete Mathematics Group, Institute for Basic Science (IBS), Daejeon, South Korea}
\icmlaffiliation{aitrics}{AITRICS, Seoul, South Korea}

\icmlcorrespondingauthor{Juho Lee}{juholee@kaist.ac.kr}
\icmlcorrespondingauthor{Hongseok Yang}{hongseok.yang@kaist.ac.kr}

\icmlkeywords{Machine Learning, ICML}

\vskip 0.3in
]



\printAffiliationsAndNotice{}  

\begin{abstract}
\input{main/abstract}

\end{abstract}

\input{main/introduction}
\input{main/backgrounds}
\input{main/methods}
\input{main/experiments}
\input{main/related_works}

\input{main/conclusion}

\section*{Acknowledgements}
This work was partially supported by Institute of Information \& communications Technology Planning \& Evaluation (IITP) grant funded by the Korea government (MSIT) (No.2019-0-00075, Artificial Intelligence Graduate School Program (KAIST)), Artificial Intelligence Innovation Hub (No.2022-0-00713), and National Research Foundation of Korea (NRF) funded by the Ministry of Education (NRF2021M3E5D9025030). HY was supported by the Engineering Research Center Program through the National Research Foundation of Korea (NRF) funded by the Korean Government MSIT (NRF-2018R1A5A1059921) and also by the Institute for Basic Science (IBS-R029-C1). We are grateful to Seongho Keum, who helped us throughout the process of building up this work.

\bibliography{references}
\bibliographystyle{icml2023}

\newpage
\appendix
\onecolumn

\input{appendix/proof}
\input{appendix/rppvsper}
\input{appendix/initializations}
\input{appendix/correlation}
\input{appendix/trajectory_comparison}
\input{appendix/details}
\input{appendix/additional}
\end{document}

%% file: preamble/preamble.tex
\usepackage{amsmath, amssymb, amsthm}
\usepackage{mathtools}
\usepackage{bbm}
\usepackage{dsfont}
\usepackage{enumitem}
\setitemize{noitemsep,topsep=0pt,parsep=0pt,partopsep=0pt}
\usepackage{wrapfig,lipsum,booktabs}

\allowdisplaybreaks

\usepackage{graphicx}
\usepackage[labelfont=bf]{caption}
\usepackage[format=hang]{subcaption}

\usepackage{booktabs, array}
\usepackage{multirow}
\usepackage{makecell}

\usepackage[colorlinks,linktoc=all,pagebackref=true]{hyperref}
\usepackage[all]{hypcap}
\hypersetup{citecolor=NavyBlue}
\hypersetup{linkcolor=NavyBlue}
\hypersetup{urlcolor=NavyBlue}
\usepackage[nameinlink,capitalise, noabbrev]{cleveref}
\creflabelformat{equation}{#2\textup{#1}#3}  
\crefname{section}{\S}{\S\S}

\input{preamble/acronyms.tex}
\input{preamble/math.tex}

%% file: preamble/acronyms.tex
\usepackage[acronym,nowarn,section,nogroupskip,nonumberlist]{glossaries}
\glsdisablehyper{}

\newcommand{\acaps}[1]{{\scshape #1}}
\newacronym[\glslongpluralkey={Projection-based Equivariance Regularizer}]{per}{\acaps{per}}{Projection-based Equivariance Regularizer}
\newacronym[\glslongpluralkey={Residual Pathway Prior}]{rpp}{\acaps{rpp}}{Residual Pathway Prior}
\newacronym[\glslongpluralkey={Soft Equivariant Multilayer Perceptrons}]{softemlp}{\acaps{s}oft\acaps{emlp}}{Soft Equivariant Multilayer Perceptrons}
\newacronym[\glslongpluralkey={Mixed EMLPs}]{memlp}{\acaps{memlp}}{Mixed EMLP}
\newacronym[\glslongpluralkey={Multilayer Perceptrons}]{mlp}{\acaps{mlp}}{Multilayer Perceptrons}
\newacronym[\glslongpluralkey={Equivariant Multilayer Perceptrons}]{emlp}{\acaps{emlp}}{Equivariant Multilayer Perceptrons}
\newacronym[\glslongpluralkey={Partial Group Convolutional Neural Networks}]{pgcnn}{\acaps{p}artial \acaps{g-cnn}}{Partial Group Convolutional Neural Networks}

%% file: preamble/math.tex
\newcommand{\ba}{\mathbf{a}}
\newcommand{\bb}{\mathbf{b}}
\newcommand{\bc}{\mathbf{c}}

\newcommand{\bx}{\mathbf{x}}
\newcommand{\by}{\mathbf{y}}
\newcommand{\bz}{\mathbf{z}}

\newcommand{\bI}{\mathbf{I}}

\newcommand{\bS}{\mathbf{S}}

\newcommand{\bV}{\mathbf{V}}

\newcommand{\calI}{{\mathcal{I}}}

\newcommand{\calN}{{\mathcal{N}}}

\newcommand{\calR}{{\mathcal{R}}}

\newcommand{\calX}{{\mathcal{X}}}
\newcommand{\calY}{{\mathcal{Y}}}

\newcommand{\bbE}{\mathbb{E}}

\newcommand{\bbR}{\mathbb{R}}

\newcommand{\balpha}{{\boldsymbol{\alpha}}}

\newcommand{\bmu}{{\boldsymbol{\mu}}}

\newcommand{\bsigma}{{\boldsymbol{\sigma}}}

\theoremstyle{plain}
\newtheorem{thm}{Theorem}[section]

\newtheorem{prop}[thm]{Proposition}

\theoremstyle{definition}

\theoremstyle{remark}

\newcommand{\normal}{\mathcal{N}}

\newcommand{\iidsim}{\overset{\mathrm{i.i.d.}}{\sim}}

\newcommand{\mathwrap}[1]{\texorpdfstring{#1}{TEXT}}

\def\[#1\]{\begin{align}#1\end{align}}

\newcommand{\norm}[1]{\lVert {#1} \rVert}
\newcommand{\gSO}{\mathrm{SO}}
\newcommand{\gS}{\mathrm{S}}
\newcommand{\gO}{\mathrm{O}}
\newcommand{\vect}{\mathrm{vec}}

\newcommand{\spm}[1]{\scriptstyle{\pm#1}}

%% file: main/abstract.tex
Datasets often have their intrinsic symmetries, and particular deep-learning models called equivariant or invariant models have been developed to exploit these symmetries. However, if some or all of these symmetries are only approximate, which frequently happens in practice, these models may be suboptimal due to the architectural restrictions imposed on them. We tackle this issue of approximate symmetries in a setup where symmetries are mixed, i.e., they are symmetries of not single but multiple different types and the degree of approximation varies across these types. Instead of proposing a new architectural restriction as in most of the previous approaches, we present a regularizer-based method for building a model for a dataset with mixed approximate symmetries. The key component of our method is what we call equivariance regularizer for a given type of symmetries, which measures how much a model is equivariant with respect to the symmetries of the type. Our method is trained with these regularizers, one per each symmetry type, and the strength of the regularizers is automatically tuned during training, leading to the discovery of the approximation levels of some candidate symmetry types without explicit supervision. Using synthetic function approximation and motion forecasting tasks, we demonstrate that our method achieves better accuracy than prior approaches while discovering the approximate symmetry levels correctly.

%% file: main/introduction.tex
\section{Introduction}
\label{main:sec:introduction}
Exploiting symmetries in a dataset is one of the key principles for building an effective deep-learning model. A popular approach for implementing this principle is to restrict the architecture of a neural network in the model so that the model has desired symmetries by construction. The approach has been highly successful, leading to a range of effective so-called equivariant or invariant models~\citep{bronstein2021geometric}, such as CNNs~\citep{cohen2016group,cohen2019gauge,choen2018spherical} and GNNs~\citep{kipf2016semi,velickovic2018graph}, that cover different types of symmetries, such as translation invariance.

\begin{figure}
     \centering
     \begin{subfigure}[b]{0.23\textwidth}
         \centering
         \includegraphics[width=\textwidth]{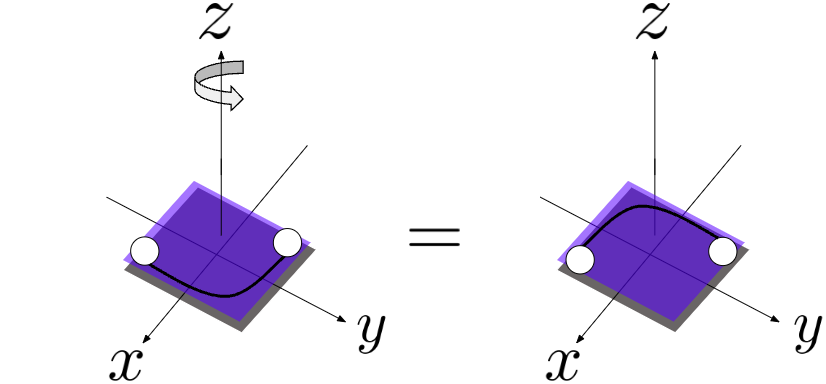}
         \caption{Equivariant trajectory}
         \label{fig:breaking1}
     \end{subfigure}
     \hfill
     \begin{subfigure}[b]{0.23\textwidth}
         \centering
         \includegraphics[width=\textwidth]{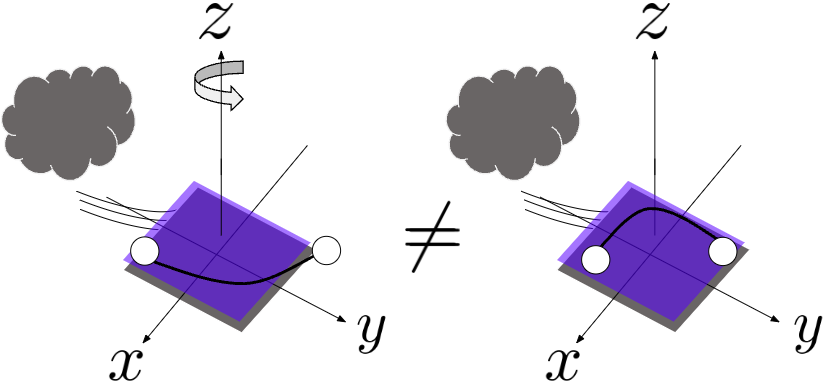}
         \caption{Soft-equivariant trajectory}
         \label{fig:breaking2}
     \end{subfigure}
        \caption{Illustrative example of a system with mixed symmetries with soft equivariances.}
        \label{fig:breaking}
\end{figure}

In practice, however, the symmetries implied in data are often approximate, partially due to measurement noises or unexpected external effects. For such scenarios, models that are equivariant or invariant by construction may be suboptimal due to their architectural restrictions. Moreover, while most of the previous works assume a single type of symmetry, many real-world data come with mixed symmetries, that is, multiple types of symmetries may exist in data. Equivariant models assuming symmetries of just a single type cannot easily be combined to model such mixed symmetries. Even more, those mixed symmetries may be approximate, so different types of symmetries may exhibit different approximation levels. 
As an example, imagine we want to model the trajectory of a golf ball in 3D space as in \cref{fig:breaking}. The trajectory is $\gO(3)$ equivariant, or there are mixed symmetries w.r.t. $\gO^x(2)$, $\gO^y(2)$, and $\gO^z(2)$. Now assume that a wind is blowing along the $y$-axis. While the trajectory is still $\gO^y(2)$-equivariant, it is only approximately equivariant to $\gO^x(2)$ and $\gO^z(2)$. An $\gO(3)$-equivariant model by design would be too restrictive in this case, and a model equivariant only with respect to $\gO^y(2)$ would miss soft equivariance along $x$ and $z$ axes.


In this paper, we tackle the modeling problem under mixed approximate symmetries, i.e., there are multiple types of symmetries with varying degrees of approximations across the types. Instead of building models symmetric by design, we propose a \emph{regularizer-based} method, where an unconstrained model is regularized toward equivariance. The regularizer is attached for each potential symmetry type expected to be implied in data, and the degree of equivariance approximation of the type is captured by the strength of the regularizer for it - its regularization coefficient. Since it is almost impossible to know the degree of approximations in advance, the regularization coefficients must be carefully tuned to capture the approximation levels correctly. Our method, without explicit supervision, can automatically tune the coefficients during training and, thus, automatically discover the varying degrees of equivariance approximations (from prescribed candidate groups) in the mixed-symmetry settings.

We are not the first to study approximate symmetries. However, the existing works mostly rely on architectural restrictions in relaxed forms~\citep{finzi2021residual,ouderaa2022relaxing,wang2022approximately}. Moreover, they do not consider multiple types of symmetries with different approximation levels. In contrast, our method does not impose architectural restrictions on a model but solely relies on the equivariance regularizers. As we will show later, the regularizer-based method is especially useful in the mixed-symmetry settings, while the existing works are not straightforwardly extended to those settings.  

We experimentally evaluated our method with a synthetic function-approximation task and a motion forecasting task. Our method could correctly discover degrees of approximations of different symmetry types in a relative term and achieve better test accuracy.

We summarize our contributions below:
\begin{itemize}
    \item We tackle the problem where we have multiple types of (approximate) symmetries with different levels of equivariance/invariance errors.
    \item We propose a novel method regularizing an unrestricted model with (approximate) symmetry constraints, and present an algorithm that can 
    automatically identify approximation levels of different symmetry types during training.
    \item We demonstrate the effectiveness of our approach on synthetic and real-world tasks with multiple types of (approximate) symmetries.
\end{itemize}

%% file: main/backgrounds.tex
\section{Backgrounds}
\label{main:sec:backgrounds}

We start with a review on the formalization of symmetries of neural networks in terms of groups. We also review so called residual pathway prior \citep{finzi2021residual}, a recent proposal for handling approximate symmetries.

\subsection{Group Representation and Equivariance}

A \emph{representation} of a group $G$ on a Euclidean space $\mathbb{R}^n$ is a function $\rho$ from $G$ to the general linear group on $\mathbb{R}^n$ (i.e., the group of invertible $n\times n$ matrices with matrix multiplication as group composition) such that $\rho$ preserves the composition operator of the group. When we have representations of a group $G$ in Euclidean spaces $\calX$ and $\calY$, denoted $\rho_\calX$ and $\rho_\calY$, we say that a function $f : \calX \to \calY$ is \emph{$G$-equivariant} if for all $g \in G$ and $x \in \calX$, we have
\[
f\big(\rho_{\cal{X}}(g)(x)\big) = \rho_{\cal{Y}}(g)\big(f(x)\big).
\]
Intuitively, this condition means that $f$ does not actively use information that can be altered by group elements $g$. The convolution layers in CNNs are a leading example that is equivariant with respect to the translation group (in an ideal setup where the images are defined over the entire plane $\mathbb{R}^2$). A range of neural-network architectures that ensure equivariance (including equivariant multilinear perceptions to be explained next) have been developed because they usually generalize better than non-equivariant counterparts.

\subsection{Equivariant Multilayer Perceptrons}

\glspl{emlp}~\citep{finzi2021practical} are models that are guaranteed to be equivariant with respect to a given group $G$ and its representation $\rho$. As its name indicates, an \gls{emlp} is identical to a standard multilayer perceptron except for one thing: its weights and biases are not network parameters, but they are constructed out of other parameters. This further parameterization of weights and biases ensure that all the linear layers of the \gls{emlp} are equivariant by construction. 

To describe the linear layers of an \gls{emlp} formally, we need to recall a few facts. First, when $G$ has representations $\rho$ on $\mathbb{R}^n$ and $\rho'$ on $\mathbb{R}^m$, the set of $G$-equivariant linear maps from $\mathbb{R}^n$ to $\mathbb{R}^m$ forms a vector space. Thus, it has an orthonormal basis $\mathcal{B} = \{M_1,\ldots,M_d\}$ where the $M_i$'s are $m\times n$ matrices representing $G$-equivariant linear maps and when reshaped to vectors via stacking columns (i.e., $\vect(M_1),\ldots,\vect(M_d)$), the matrices become orthonormal vectors of $(m \times n)$ dimension. Second, the set of vectors $v$ in $\mathbb{R}^m$ that are invariant with respect to $G$ and $\rho'$ (i.e., $\rho'(g)(v) = v$ for all $g \in G$) forms a subspace of $\mathbb{R}^m$. So, this subspace also has an orthonormal basis. The linear layers of \gls{emlp} are defined in terms of these two bases.

Assume that the $l$-th layer of the network has $n_l$ input nodes and $n_{l+1}$ output nodes. Formally, each linear layer $l$ of an \gls{emlp} is an affine map
$\mathrm{Linear}_\text{EMLP} : \mathbb{R}^{n_l} \to \mathbb{R}^{n_{l+1}}$
defined as follows:
\[
\label{eq:definition-emlp}
\lefteqn{\mathrm{Linear}_\text{EMLP}(x) = Wx+b,}\nonumber\\
&\mathrm{vec}(W) = Q\theta, \quad b = R\beta,
\]
where $\mathrm{vec}(W)$ is the vector obtained by stacking the columns of the matrix $W$, $Q$ is a fixed matrix with $(n_{l+1} \times n_l)$ rows and $d$ columns, and $R$ is a fixed matrix with $n_{l+1}$ rows and $r$ columns. The columns of the matrix $Q$ when reshaped to $n_{l+1} \times n_l$ matrices via unstacking form an orthonormal basis of the space of $G$-equivariant linear maps from $\mathbb{R}^{n_l}$ to $\mathbb{R}^{n_{l+1}}$. Similarly, the columns of the other matrix $R$ form the basis of the subspace of $G$-invariant vectors in $\mathbb{R}^{n_{l+1}}$. The parameters to be trained are $\theta \in \bbR^{d}$ and $\beta \in \bbR^{r}$, the coefficients combining the orthonomal basis.




\subsection{Residual Pathway Prior}

The \gls{rpp} \citep{finzi2021residual} is a recent proposal for learning an approximately-equivariant neural network. It is based on the idea of combining equivariant and non-equivariant transformations together in each network layer. Concretely, it is the following variant of the \gls{emlp}, which adds a standard linear layer, called residual pathway, to each equivariant linear layer of the \gls{emlp}:
\[
\label{eq:linear-rpp}
\lefteqn{\mathrm{Linear}_\text{RPP}(x) = Wx + b,}\nonumber\\
&\mathrm{vec}(W) = QQ^\intercal \mathrm{vec}(W_1) + \mathrm{vec}(W_2),\nonumber\\
&b = RR^\intercal b_1 + b_2,
\]
where $Q$ and $R$ are from the equations in \eqref{eq:definition-emlp}, and $Q^\intercal \mathrm{vec}(W_1)$ and $R^\intercal b_1$ correspond to
$\theta$ and $\beta$ in the same equations, respectively. Note that $\mathrm{Linear}_\text{RPP}(x)$ is the sum of the \gls{emlp}'s linear layer on $x$ and 
$W_2x + b_2$. The residual pathway refers to the latter part.

The parameters of an \gls{rpp} are trained with the following $\ell_2$-regularization, which comes from the prior distributions on those parameters:
\[
\begin{split}
\label{eq:regularizer-rpp}
\lefteqn{\calR^\text{RPP}(W_1,b_1,W_2,b_2)}\\
&= \frac{\norm{\mathrm{vec}(W_1)}^2+\norm{b_1}^2}{2\sigma_1^2}+\frac{\norm{\mathrm{vec}(W_2)}^2+\norm{b_2}^2}{2\sigma_2^2},
\end{split}
\]
with $\sigma_2$ being substantially smaller than $\sigma_1$, which encourages that residual layers play only a minor role for inference. 

%% file: main/methods.tex
\section{Equivariance Regularizer}
\label{main:sec:methods}

In this section, we present our equivariance regularizer, the key conceptual contribution of the paper. We assume that a collection of groups $G_1,G_2,\ldots,G_K$ are given, capturing different types of symmetries, and also that these groups come with representations for input and output spaces of all linear layers. The latter assumption enables us to talk about $G_k$-equivariant linear or affine maps for all layers. In our presentation, we fix a layer $l$, and describe how our regularizers constrain network parameters at that layer. For notational simplicity, we omit the layer indices from the parameters,
unless required to be specified.

\subsection{Projection-Based Equivariance Regularizer}
\begin{figure}
    \centering
    \includegraphics[width = 0.32\textwidth]{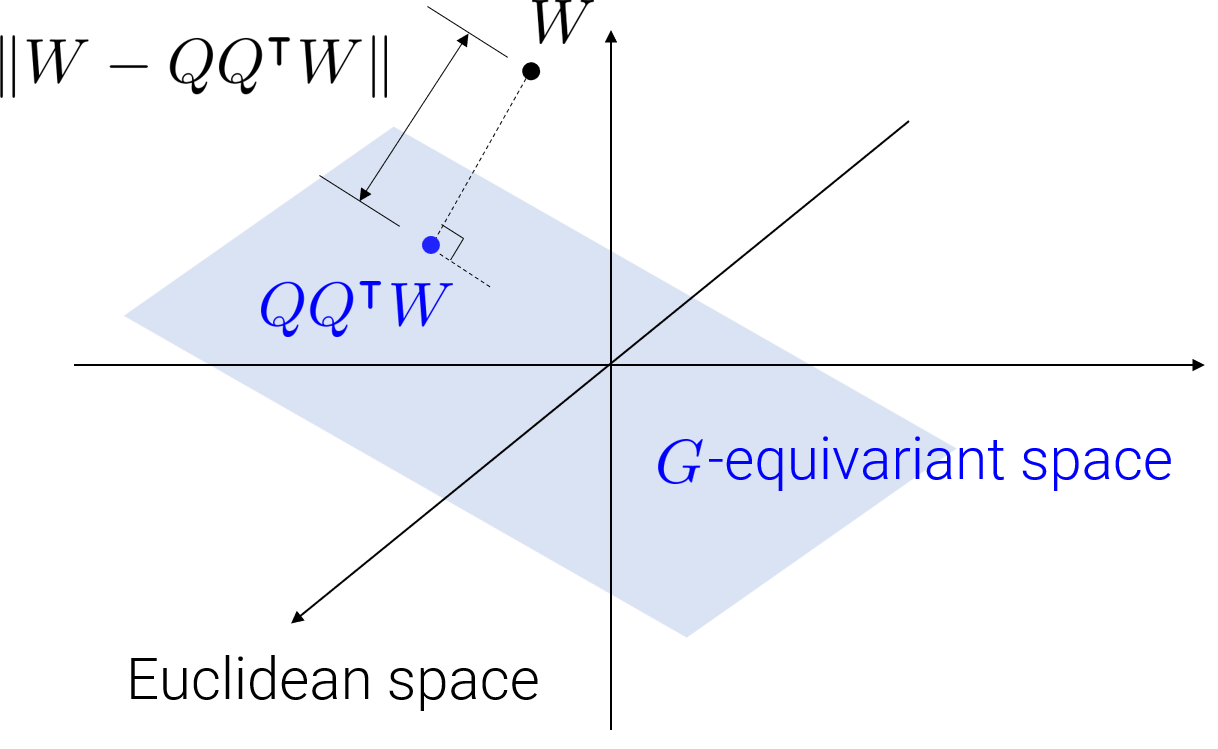}
    \caption{The projection-based equivariance regularizer for a group $G$ measures the distance $\norm{W-QQ^\intercal W}$, where $W$ is either $\mathrm{vec}(W)$ or $b$ in the standard linear layer and $Q$ is an orthonormal basis of the space of $G$-equivariant matrices or $G$-invariant vectors.}
    \label{fig:per}
\end{figure}

For every $k = 1,\ldots,K$, write $Q_k$ and $R_k$ for the matrices from the equations in \eqref{eq:definition-emlp}; the columns of $Q_k$ form an orthonormal basis of $G_k$-equivariant linear maps from $\mathbb{R}^{n_l}$ to $\mathbb{R}^{n_{l+1}}$ after being reshaped into $n_{l+1}\times n_{l}$ matrices, and the columns of $R_k$ form an orthonormal basis of $n_{l+1}$-dimensional $G_k$-invariant vectors in $\mathbb{R}^{n_{l+1}}$. 

Our \gls{per} for a group $G_k$ is defined by
\[
\label{eq:per}
\calR^\text{PER}_k(W,b)
&= \frac{\lambda_k}{2}\norm{\mathrm{vec}(W)-Q_kQ_k^{\intercal}\mathrm{vec}(W)}^2 \nonumber\\
& + \frac{\lambda_k}{2}\norm{b-R_kR_k^{\intercal} b}^2,
\] where $W$ and $b$ are parameters of the $l$-th layer of the network, and $\lambda_k$ is a regularization coefficient for the group $G_k$. Modulo the reshaping into the vector form, the term $Q_kQ_k^{\intercal}\mathrm{vec}(W)$ is the projection of $W$ (expressing a linear map from $\mathbb{R}^{n_l}$ to $\mathbb{R}^{n_{l+1}}$) to the space of $G_k$-equivariant linear maps expressed as $n_{l+l} \times n_l$ matrices. Thus, the first summand measures the $\ell_2$-distance from $W$ to the space of $G_k$-equivariant linear maps. Similarly, the second summand uses the projection of the bias term and measures the $\ell_2$-distance from $b$ to the space of $G_k$-invariant vectors. This regularizer can be a part of a learning objective during training, so that the training moves the parameters $W$ and $b$ towards the space of the $G_k$-equivariant linear maps or $G_k$-invariant vectors. An advantage of this regularizer-based approach for enforcing symmetries is that we can easily combine multiple regularizers for different groups simply by adding them to the objective function. Concretely, in our setup of $K$ different groups, we can use the following regularizer for the parameters of the $l$-th layer:
\[
\label{eq:multi-per}
\calR^\text{PER}(W,b) = \sum_{k=1}^K \calR_k^\text{PER}(W, b)
\]
The regularization coefficients $\lambda_k$ control the strength of enforcing different types of symmetries formalized by different groups $G_1,\ldots,G_K$. Ideally, these parameters are set according to the approximation levels of different symmetry types. However, we don't know the approximation levels in advance. In the next section, we explain how to infer such parameters during training without explicit supervision.

An implicit assumption under the regularizer $\calR^\text{PER}(W, b)$ is that the $\ell_2$-distance measures how much the symmetry with respect to $G_k$ is violated by the corresponding parameters of the network. The following proposition supports that assumption, showing that minimizing the $\ell_2$ distances indeed minimizes the equivariance error.

\begin{prop}[]
\label{prop:inequality}
Let $f$ be an $S$-layer MLP with the weight matrix $W^{(l)}$ and the bias term $b^{(l)}$ at each layer $l$. Assume that the activation functions of $f$ are $G$-equivariant and $L$-Lipchitz continuous. Also, assume a constant $U > 0$ such that 
$\|x\| < U$ for every $x \in \calX$, and the operator norms  $\norm{\rho_\calX(g)}_\mathrm{op}$
and $\norm{\rho_\calY(g)}_\mathrm{op}$
for any $g \in G$ are also bounded by $U$. Then, there exists a constant $C > 0$ depending on $S$, $L$, and $U$ only, such that for all $\{(W^{(l)},b^{(l)})\}_{l = 1,\ldots,S}$,
if the operator norm $\norm{W^{(l)}}_\mathrm{op}$ and the $\ell_2$ norm $\norm{b^{(l)}}$ are bounded by $U$ for every $l$, we have
\begin{align}
\nonumber
& 
\sup_{x\in\calX,g\in G}\norm{\rho_\calY(g)f(x)-f(\rho_\calX(g)x)}
\\[-1ex]
\nonumber
&
\quad {} \leq
C \cdot \sum_{l=1}^S\bigg(\norm{\mathrm{vec}(W^{(l)})-Q_k^{(l)}Q_k^{(l)\intercal}\mathrm{vec}(W^{(l)})}
\\[-1ex]
&
\qquad\qquad\qquad {} +\norm{b^{(l)}-R_k^{(l)}R_k^{(l)\intercal} b^{(l)}}\bigg).
\end{align}
\end{prop}
The proof of a refined version of this proposition is given in \cref{app:sec:proof}. According to \cref{prop:inequality}, the equivariance error of a model is bounded by the $\ell_2$-distances of parameters to equivariance subspaces, and the minimum equivariant error is achieved when the $\ell_2$-distances are zero, which happens when the value of the regularizer is zero.

Our methodology presents a comparable functionality to \gls{rpp}, yet it allows the model to discover soft equivariant weights with a reduced number of parameters. The distinctions between \gls{rpp} and \gls{per} have been visually depicted in~\cref{app:sec:compare}.

\subsection{Adjustment of Hyperparameters of Groupwise Equivariance Regularizers}
\label{main:sec:autotune}

The regularization coefficients $\lambda_1,\dots, \lambda_K$ in \eqref{eq:multi-per} play an important role of controlling the strengths of groupwise equivariance constraints that we impose on the model. We empirically observed that a better model is learned when these regularization coefficients for different groups (and hence the strengths of regularization for these groups) are correlated with the approximation levels of symmetries for those groups in a dataset. That is, if $(\lambda_1^*, \dots, \lambda_K^*)$ are the coefficents leading to the best model with the lowest validation error after training, then a smaller $\lambda_k^*$ value means weaker symmetry (more approximation error) for the group $G_k$, and a larger $\lambda_k^*$ means more exact symmetry for the group $G_k$. 

Based on this observation, we propose an automatic tuning procedure that could discover the approximation levels of different symmetry types (formalized by different groups and captured through the regularizers) in a data-driven way. Given an $S$-layer MLP $f$, 
let $\calR_k^\text{PER}(f) = \sum_{\ell=1}^S\calR^\text{PER}_k(W^{(l)}, b^{(l)})$. We first initialize all the regularization coefficients with the same value, and in the early stage of training, adjust the coefficients $\{\lambda_k\}_{k=1,\ldots,K}$ based on the magnitudes of the corresponding regularizers $\{\calR^\text{PER}_{k}(f)\}_{k=1,\ldots,K}$ with the following formula:
\[
\lambda_k^* = \lambda_k\Bigg(\frac{\min\{\calR_k^\text{PER}(f)\}_{k=1,\ldots,K}}{\calR^\text{PER}_{k}(f)}\Bigg)^\gamma,
\] where $\gamma$ is a scaling factor calibrating how much the approximation difference will be reflected in the coefficients. We empirically confirmed that setting $\gamma \in [2, 5]$ gives reasonable results.

\subsection{Extension of \gls{emlp} for Mixed Symmetries}
\label{main:sec:methods:mixedemlp}
Unlike our method which can conveniently combine multiple regularizers for mixed symmetries, it is not straightforward to extend existing (approximately) equivariant models for mixed symmetry settings. Here, as a baseline, we describe a na\"ive extension of \gls{emlp} for our setup which assumes multiple types of symmetries formalized by groups $G_1,\ldots,G_K$. Assume the model is equivariant to the first $L$ groups $G_1, \dots, G_L$ and softly equivariant for the rest. For $G_1,\dots, G_L$, we first compute a joint subspace by solving the set of equivariance constraints for $L$ groups and denote the corresponding bases $Q_1$ and $R_1$. Similarly, we compute a joint subspace for all groups $G_1, \dots, G_K$ and denote the bases $Q_2$ and $R_2$. A \gls{memlp} is defined as
\[\label{eq:linear-mixedemlp}
&\mathrm{Linear}_\text{MEMLP}(x) = W_1x + b_1 + W_2x + b_2\, \nonumber\\
& \mathrm{vec}(W_q) = Q_q\theta_q, \,\,\, b = R_q\beta_q \text{ for } q=1,2.
\]
Here, both $W_1x+b_1$ and $W_2x+b_2$ are equivariant to $G_1,\dots, G_L$, so the overall model is equivariant to them. On the other hand, since $W_1x+b_1$ is not equivariant to $G_{L+1},\dots, G_K$, the overall model is only softly equivariant to them.  The level of soft equivariance is controlled by the prior variances for $W_1$ and $W_2$, as in the case of \gls{rpp}. 



%% file: main/experiments.tex
\section{Experiments}
\label{main:sec:experiments}
\glsunset{memlp}
To demonstrate the effectiveness of our method, especially for its utility in discovering mixed symmetries from data, we compare ours to (approximately) equivariant baseline models for a synthetic function approximation task and a real-world motion forecasting task. The baselines we are comparing against include \gls{emlp}, \gls{rpp}, and \gls{memlp} described in~\cref{main:sec:methods:mixedemlp}. The network architectures used for those models including our model in common have four layers with the gated nonlinearities and bilinear layers as described in \citet{finzi2021practical,finzi2021residual}.
Throughout all the experiments, to see the net effect of the abilities of the models capturing equivariances, we controlled the sizes of the competing models so that all of them have similar number of parameters.

Additional information regarding the experiments, such as the specific hyperparameters employed and the data preprocessing details applied, can be found in~\cref{app:sec:details}. Furthermore, \cref{app:sec:initialization} provides insightful recommendations for efficient initializations of neural networks in the \gls{per} settings. Additionally, \cref{app:sec:additional} presents supplementary experiments conducted to assess the robustness of our method.


\subsection{Synthetic Function-Approximation Task}

\input{tables/testmse_inertia}
\subsubsection{The moment of Inertia Function}
We generate a synthetic dataset having mixed symmetries by adding a perturbation to a symmetric function that computes the moment of inertia. Given the masses and positions of five particles, denoted by $(m_{1:5}, \bx_{1:5}) := (m_i,\bx_i)_{i=1}^5$, the moment of inertia is computed as follows:
\[
\label{eq:moment-inertia}
\calI(m_{1:5}, \bx_{1:5}) := \sum_{i=1}^5m_i(\bx_i^\intercal\bx_i\bI-\bx_i\bx_i^\intercal).
\]
The moment-of-inertia function is equivariant with respect to group $\gO(3)$, which consists of rotations and reflections. That is, for a group element $g\in\gO(3)$, $\rho(g)\calI(m_{1:5}, \bx_{1:5})=\calI(\rho(g)(m_{1:5}, \bx_{1:5}))$. Here $g$ acts on each position $\bx_i$, that is, $\rho(g)(m_{1:5}, \bx_{1:5}) = (m_i,g\bx_i)_{i=1}^5$ where $g$ in $g\bx_i$ is represented as a $3\times 3$ matrix. The output of the function $M = \calI(m_{1:5}, \bx_{1:5})$
is a $3\times 3$ matrix, and $g$ acts on $M$ as $\rho(g)(M) = gMg^{-1}$.

To generate data, we draw $\bx_{1:5} \iidsim \calN(\mathbf{0}, \bI)$ and $m'_{1:5} \iidsim \calN(0, 1)$, and then compute $m_i = \mathrm{softplus}(m_i')$. We then compute the moment of inertia with \eqref{eq:moment-inertia} and add five different types of errors to the output. Let $\hat\bx, \hat\by, \hat\bz \in \bbR^3$ be the orthonormal basis vectors of the $x, y$ and $z$ axes, respectively. The five types of errors and the corresponding approximate symmetries are as follows:
\begin{enumerate}[topsep=0pt,itemsep=-0.5ex,partopsep=1ex,parsep=1ex]
    \item $\mathbf{0}$ (no error), $\gO(3)$-equivariant.
    \item $-\calI\hat{\bx}\hat{\bx}^\intercal$, $\gO^x(2)$ equivariant, soft $\gO(3)$ equivariant.
    \item\label{item:intertia_error_3} $-\calI\hat\by\hat\by^\intercal$, $\gO^y(2)$ equivariant, soft $\gO(3)$-equivariant.
    \item\label{item:intertia_error_4} $-\calI\hat\bz\hat\bz^\intercal$, $\gO^z(2)$ equivariant, soft $\gO(3)$ equivariant.
    \item $-0.3\calI(\hat\bx\hat\bx^\intercal -\hat\by\hat\by^\intercal + \hat\bz\hat\bz^\intercal)$, soft $\gO(3)$-equivariant.
\end{enumerate}
For the baselines, we consider $\gO(3)$\gls{emlp}, $\gO(3)$\gls{rpp}, and $\gO^{\text{(axis)}}$-$\gO(3)$\gls{emlp} which is equivariant to $\gO^{\text{(axis)}}$ and softly equivariant to $\gO(3)$, where $\text{axis} \in \{x, y, z\}$ is chosen according to the symmetry in the data. Our model, denoted by \gls{per}, regularizes an MLP with equivariance regularizers for the groups $(\gO^x(2), \gO^y(2), \gO^z(2))$.

\input{tables/testmse_cossim}

\subsubsection{The CosSim Function}
Another synthetic function-approximation task we consider is the CosSim function which computes the average cosine similarity between three particles. Given the positions of three particles in 3D space, denoted by $\bx_{1:3} := \{\bx_i\}_{i=1}^3$ with each $\bx_i \in \bbR^3$, the CosSim function computes
\[
\label{eq:avgcos}
\lefteqn{\mathrm{AvgCS}(\bx_{1:3})}\nonumber\\
&= \frac{\mathrm{CS}(\bx_1,\bx_2)+\mathrm{CS}(\bx_2,\bx_3)+\mathrm{CS}(\bx_1,\bx_3)}{3},
\]
where $\mathrm{CS}(\ba,\bb) := \frac{\ba\cdot\bb}{\norm{\ba}\norm{\bb}}$. The $\mathrm{AvgCS}$ function is invariant to both $\gSO(3)$ and $\gS(3)$ where $\gSO(3)$ is a rotation group in $\bbR^3$ and $\gS(3)$ is a scaling group in $\bbR^3$. That is, for a group element $g\in \gSO(3)$ or $g\in\gS(3)$, $\mathrm{AvgCS}(\rho(g)\bx_{1:3})=\mathrm{AvgCS}(\bx_{1:3})$, where $\rho(g)\bx_{1:3}=\{g\bx_i\}_{i=1}^3$. Similarly to the inertia task, to generate data, we draw $\bx_{1:3} \iidsim \calN(\mathbf{0}, \bI)$, compute \eqref{eq:avgcos}, and inject four different types of errors.
\begin{enumerate}[topsep=0pt,itemsep=-0.5ex,partopsep=1ex,parsep=1ex]
    \item $\mathbf{0}$ (no error), $\gSO(3)$ and $\gS(3)$ invariant.
    \item $ \frac{-\sum_{i=1}^3\norm{\bx_i}}{3}$, $\gSO(3)$-invariant, soft $\gS(3)$-invariant.
    \item $\frac{-\sum_{i=1}^3 |\bx_i\cdot \hat{\bx}|}{\sum_{j=1}^3(|\bx_j\cdot\hat\by| + |\bx_j\cdot\hat\bz|)}$, soft $\gSO(3)$ invariant, $\gS(3)$ invariant.
    \item $ \frac{-\sum_{i=1}^3\norm{\bx_i}}{3}+\frac{\sum_{i=1}^3 |\bx_i\cdot \hat{\bx}|}{\sum_{j=1}^3(|\bx_j\cdot\hat\by| + |\bx_j\cdot\hat\bz|)}$, soft $\gSO(3)$ and $\gS(3)$ invariant.
\end{enumerate}
For the baselines, we consider $(\gSO(3),\gS(3))$\gls{emlp}, $(\gSO(3),\gS(3))$\gls{rpp}, 
$\gSO(3)$-$\gS(3)$\gls{memlp} (equivalent to $\gSO(3)$ and softly equivalent to $\gS(3)$), and 
$\gS(3)$-$\gSO(3)$\gls{memlp} (equiv to $\gS(3)$ and softly equiv to $\gSO(3)$). For computing the basis of the joint equivariant subspace of $\gSO(3)$ and $\gS(3)$, we solve for the conjunction of equivariance constraints for two groups, as we explained in \cref{main:sec:methods:mixedemlp}.

\subsubsection{Analysis of the results}
\begin{figure}[t]
 \centering
     \includegraphics[width=0.4\textwidth]{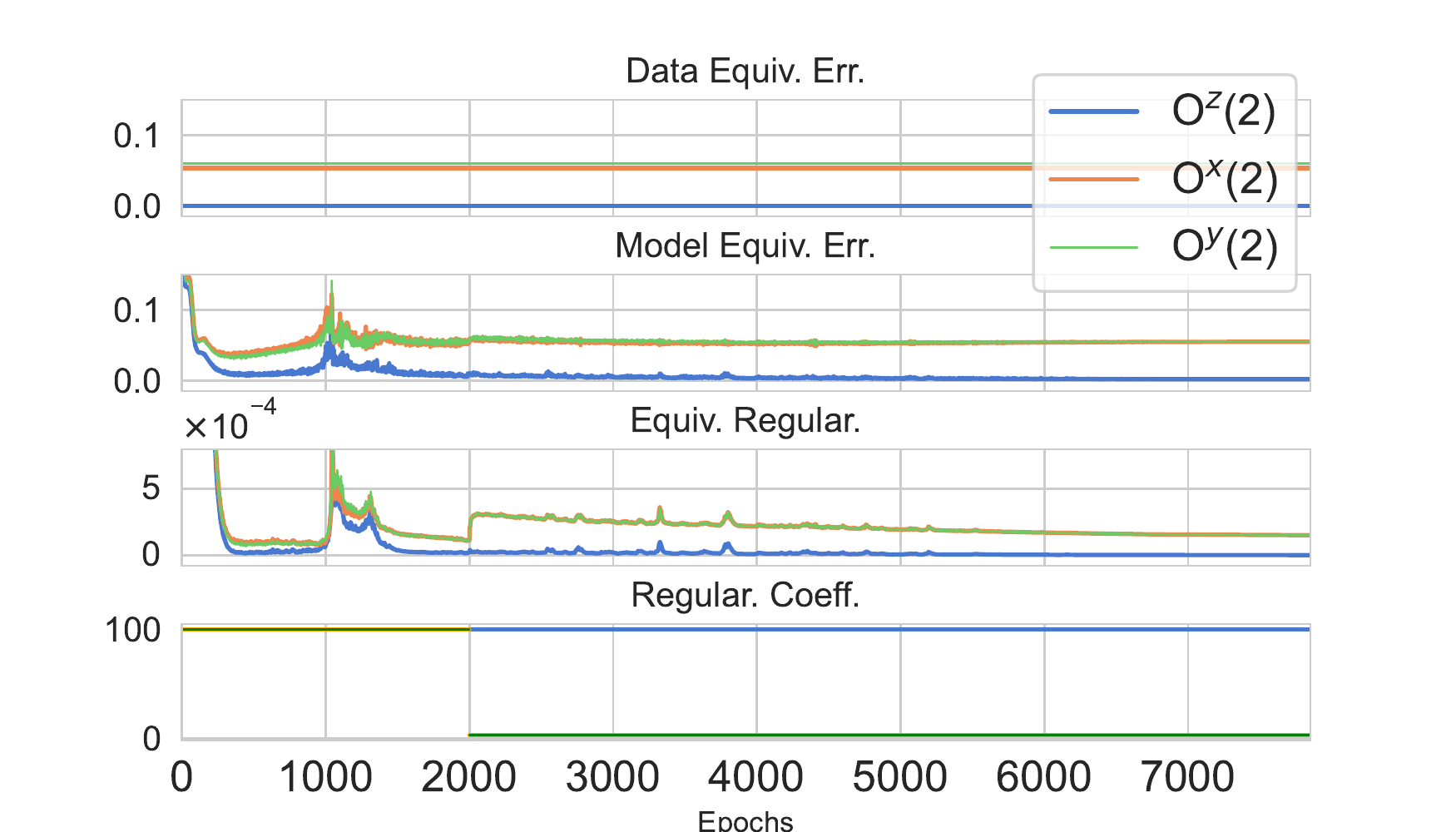}
     \caption{
     The training progress of \gls{per} on a dataset equivariant to $\gO^z(2)$ and softly equivariant to $\gO^x(2)$ and $\gO^y(2)$.
     From top to bottom: data equivariance error (defined in \cref{eq:approx-equiv-error}), model equivariance error, the values of the equivariance regularizers $(\calR^\text{PER}_k(f))_{k=1}^3$, and the regularization coefficients $(\lambda_k)_{k=1}^3$. The coefficients are adjusted automatically at epoch 2000.}     
     \label{fig:tendency}
 \end{figure}

\paragraph{Overall results.}
We summarize the results for the moment of inertia task in \cref{tab:test-inertia} and the results for the CosSim task in \cref{tab:test-cossim}. For both tasks, \gls{per} significantly outperforms baselines, across all error types having different types of approximate equivariance. From below, we empirically show that this is because \gls{per} correctly captures the approximate equivariance in data and adjusts the regularization coefficients accordingly.

\paragraph{Discovery of approximate equivariance.}
We check whether our model correctly learns the degree of approximate equivariance implied in the dataset. For instance, in the moment of inertia task, when data is perturbed by the \cref{item:intertia_error_3}, our model should be able to detect that the data is $\gO^y(2)$ equivariant and softly $\gO(3)$ equivariant. 
\cref{fig:tendency} illustrates the progress of equivariance errors, values of regularization, and their coefficients during training. Here, the data is perturbed by the error type \cref{item:intertia_error_4}, so it is $\gO^z(2)$ equivariant and softly equivariant to $\gO^x(2)$ and $\gO^y(2)$. As we can see in the figure, our model captures the difference between the equivariance error levels and adjusts the regularization coefficients at the epoch 2000. Here, our model lowers the regularization coefficients for $\gO^x(2)$ (to 2.91) and $\gO^y(2)$ (to 3.07) while keeping the coefficient for $\gO^z(2)$ (to 100.0). As a result, the model trained with the adjusted regularization could correctly match the equivariance errors assumed in the data. 

 \begin{figure}[t]
     \centering
     \includegraphics[width=0.38\textwidth]{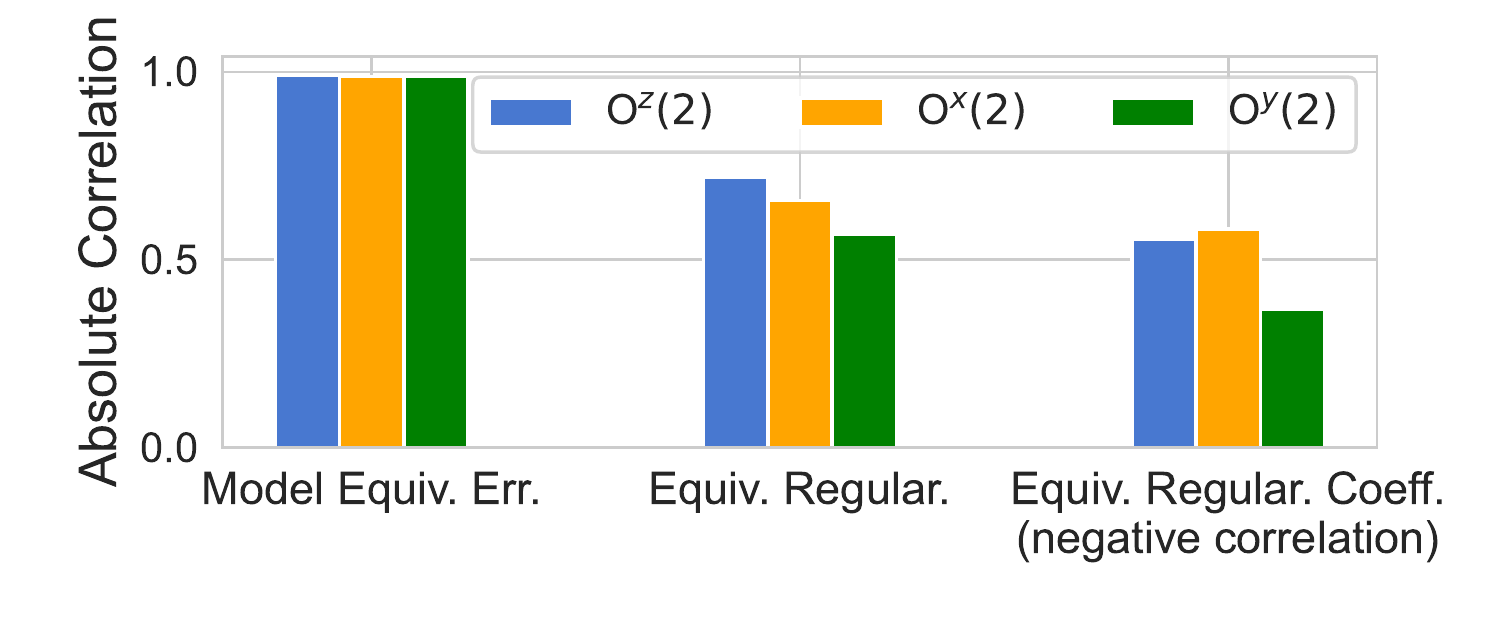}
     \caption{ Absolute Pearson correlation coefficients between data equivariance errors and (model equivariance error, the values of equivariance regularizers, the regularization coefficients), 
     measured across 13 datasets of different degrees and types of approximate equivariances.}
     \label{fig:correlation}
 \end{figure}

To further demonstrate that our model indeed captures the equivariance error levels from data, we measure the Pearson correlation coefficients between the (model equivariance errors, the values of the equivariance regularizers $(\calR_k^\text{PER}(f))_{k=1}^3$, the regularization coefficients $(\lambda_k)_{k=1}^3$), and the equivariance errors assumed in the data. Here, the model equivariance error is measured as a Monte Carlo approximation of the following expectation of equivariance error (scaled) of a model $f$:
\[
\label{eq:approx-equiv-error}
\bbE_{\bx\in\calX,g\in G}\bigg[\frac{\norm{\rho_\calY(g)f(\bx)-f(\rho_\calX(g)\bx)}}{\norm{\rho_\calY(g)f(\bx)}\norm{f(\rho_\calX(g)\bx)}}\bigg].
\]
We measure the correlations across 13 different types of datasets with varying error types and scales, and summarize the result in \cref{fig:correlation} (the specific values for each sample are written in \cref{app:sec:correlation}). The model equivariance error is highly correlated with the data equivariance error, indicating that the model correctly captures the equivariance errors implied in the data. Equivariance regularizers and their coefficients are also correlated with the data equivariance error, supporting our claim that the automatic tuning procedure in our method can discover the approximate equivariance (from prescribed candidate groups) in a data-driven way.
 


\subsection{Motion Forecasting Task}
\subsubsection{Task Description}
The goal of this task is to predict the future positions of a moving vehicle given past positions. The position of the vehicle is represented with a 3D coordinate $(x, y, z)$. We collect the trajectories from Waymo Open Motion Dataset (WOMD)~\citep{ettinger2021large} containing trajectories of vehicles moving on roads. We use 16,814 trajectories for training, 3,339 trajectories for validation, and 3,563 trajectories for testing. Each trajectory consists of $T=6$ past positions $\bx^{(1:T)}:=\{\bx^{(t)}\}_{t=1}^T$ and $T=6$ future positions $\by^{(1:T)}:=\{\by^{(t)}\}_{t=1}^T$ to be predicted, and the positions are measured at a frequency of 2.5Hz. We assess the performance of the models trained for this task using the Average Distance Error (ADE) defined as follows:
\[
\text{ADE}(\by^{(1:T)}, \hat\by^{(1:T)}) = \frac{1}{T}\sum_{t=1}^T \norm{\by^{(t)}-\hat\by^{(t)}},
\]
where $\by^{(1:T)}$ and $\hat\by^{(1:T)}$ are predicted and ground-truth future trajectories, respectively.



In principle, the trajectory of a moving vehicle is equivariant to the rotations along the $z$-axis. Therefore, an $\gO^z(2)$-equivariant model is expected to perform better than non-equivariant models. Indeed, on the WOMD dataset, \citet{assaad2022vntransformer} reported that an $\gO^z(2)$-equivariant transformer works better than a non-equivariant transformer. However, they also reported that on the same task, the $\gO^z(2)$-equivariant transformer performs \emph{worse} than a soft $\gO^z(2)$-equivariant transformer. In our experiment, we attempt to see why this is the case and also find out what other types of (approximate) symmetries the dataset might exhibit. To this end, we compare $\gO^z(2)$-\gls{emlp}, $\gO(3)$-\gls{emlp}, $\gO(3)$-\gls{rpp}, $\gO^z(2)$-\gls{rpp}, $\gO^z(2)$-$\gO(3)$ \gls{memlp}, 
and MLP with $(\gO^x(2), \gO^y(2), \gO^z(2))$ \gls{per}.


\subsubsection{Normalization Methods}


Typically, for a regression problem, we preprocess the inputs either by normalizing or scaling them. However, we find that training with trajectories with such typical preprocessing performs poorly, due to high variance across trajectories. Hence, before the actual normalization, we first do \emph{centering} for each trajectory to bring it near the origin. Given a $i$-th trajectory $\bx^{(1:T)}_i$, the centering is defined as
\[
\text{centering}(\bx_i^{(1:T)}) = (\bx_i^{(t)}-\bar\bx_i) := \bc_i^{(1:T)},
\]
where $\bar\bx_i := \sum_{t=1}^T \bx_i^{(t)}/T$.

Even after centering, we still suffer from varying scales of the coordinates (the values of the $z$-axis are significantly smaller than the values of the other axes because most vehicles run on horizontal roads). To resolve this, we may normalize each coordinate separately but it might also break the symmetry implied in the data. Hence, we consider three different types of normalization schemes where each scheme induces different (approximate) symmetry, and compare the models on the datasets preprocessed with them. The goal of the experiment is to show that our method can capture different types of symmetries induced by the normalizations and thus perform robustly across datasets. Examples of trajectory for each normalization are visually compared in~\cref{app:sec:trajectory-comparison}.

\paragraph{Scale-aware normalization.}
Assume we have $N$ trajectories in the training set.
Let $\bmu \in \bbR^3$ and $\bsigma \in \bbR_+^3$ be the element-wise mean and standard deviation of the trajectories in the training set,
\[
\bmu = \sum_{i=1}^N \sum_{t=1}^T \frac{\bc_i^{(t)}}{NT}, \,\,
\bsigma = \bigg(\sum_{i=1}^N\sum_{t=1}^T \frac{(\bc_i^{(t)}-\bmu)^{\odot 2}}{NT}\bigg)^{\odot \frac{1}{2}},
\]
where $\odot$ denotes the element-wise exponentiation. Given $\bmu$ and $\bsigma$, the first normalization scheme is defined as
\[
\text{normalize}(\bc_i^{(1:T)}) = \big((\bc_i^{(t)} - \bmu) \oslash \bsigma \big)_{t=1}^T,
\]
where $\oslash$ denotes the element-wise division. We call this normalization a \emph{scale-aware} normalization, since it adjusts the data for each coordinate separately so that all the $(x, y, z)$ coordinates have similar scales.

\paragraph{Symmetry-aware normalization.} Note that the scale-aware normalization breaks the rotation symmetry because it scales each coordinate with a different value. In that case, we may lose the benefits of utilizing the rotation equivariance in a model. In the second normalization scheme, instead of element-wise scaling, we use the total standard deviation for the scaling:
\[
&m = \sum_{i=1}^N\sum_{t=1}^T \frac{\mathbf{1}_3^\intercal\bc_i^{(t)}}{3NT}, \quad
s^2 = \sum_{i=1}^N\sum_{t=1}^T \frac{\norm{\bc_i^{(t)}-m\mathbf{1}_3}^2}{3NT} \nonumber\\
&\text{normalize}(\bc_i^{(1:T)}) = \big( (\bc_i^{(t)} - \bmu)/s \big)_{t=1}^T,
\]
where $\mathbf{1}_3=[1,1,1]^\intercal$. We call this normalization \emph{symmetry-aware} since the rotation symmetry of the resulting trajectory is not broken by the normalization.

\paragraph{Symmetry-scale-aware normalization.}
While the symmetry-aware normalization preserves the rotation symmetry, it still has the problem of a small $z$-scale in the training set. To further resolve this, as the third scheme, we modify the centering step as follows,
\[
\text{centering}(\bx_i^{(1:T)}) = (\bx_i^{(t)}- \balpha \otimes \bar\bx_i)_{t=1}^T,
\]
where $\otimes$ denotes the element-wise multiplication and $\balpha \in \bbR^3$ is a scaling factor. We set $\balpha = (1, 1, 0.993)$, so the values for the $z$-axis remain similar to the other axes after centering. Then we normalize the centered data as in the symmetry-aware normalization. Since the values of the $z$-axis were similar to those of other axes, even after the scaling, the values of the three coordinates have a similar scale. We call this scheme \emph{symmetry-scale-aware} since it is both scale-aware and preserves rotation symmetry.

\subsubsection{Analysis of the results}

 \begin{figure}[t]
     \centering
     \includegraphics[width=0.44\textwidth]{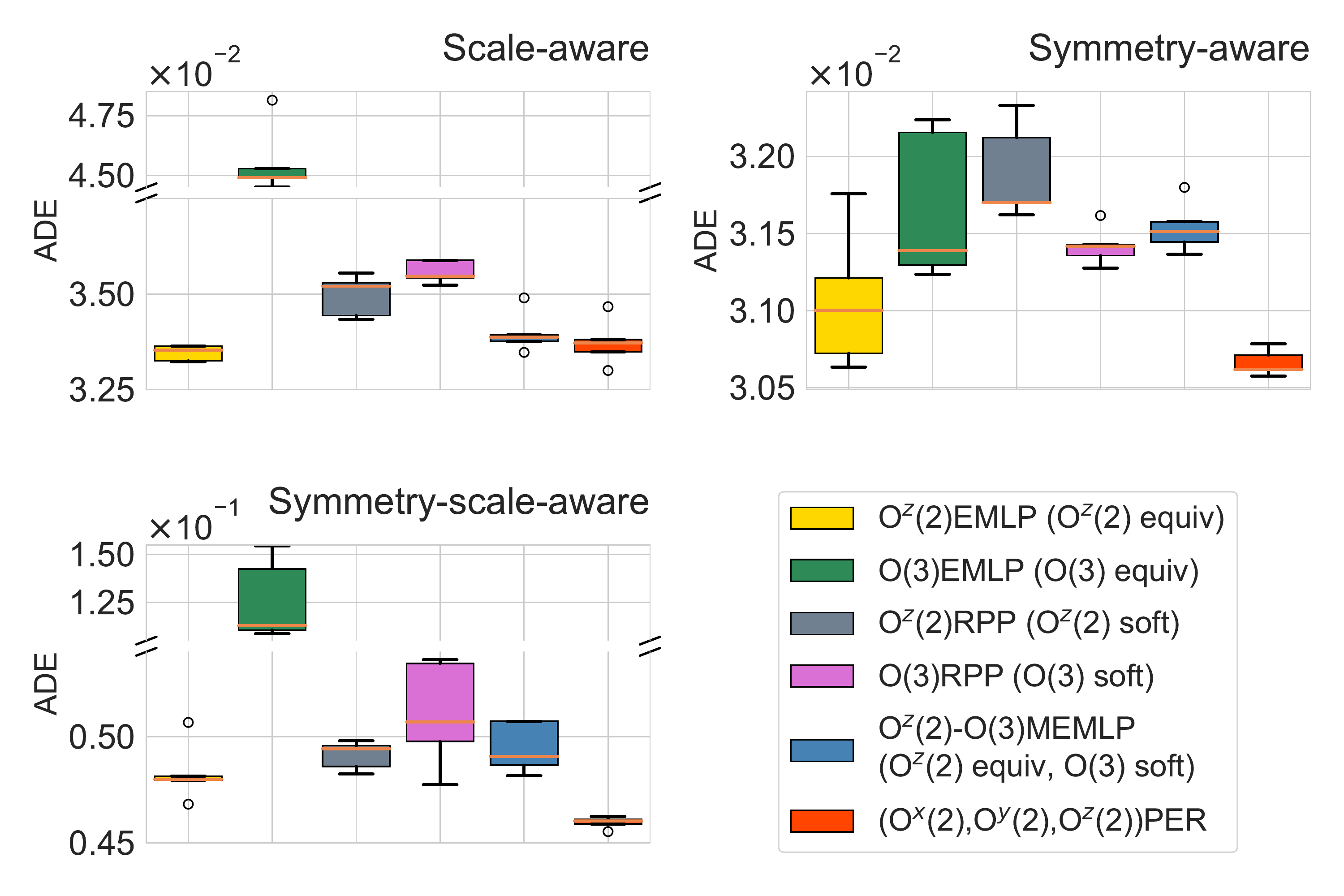}
     \caption{Test ADE results for WOMD dataset.}
     \label{fig:test-motion}
 \end{figure}

\begin{figure}[t]
    \centering
    \includegraphics[width = 0.45\textwidth]{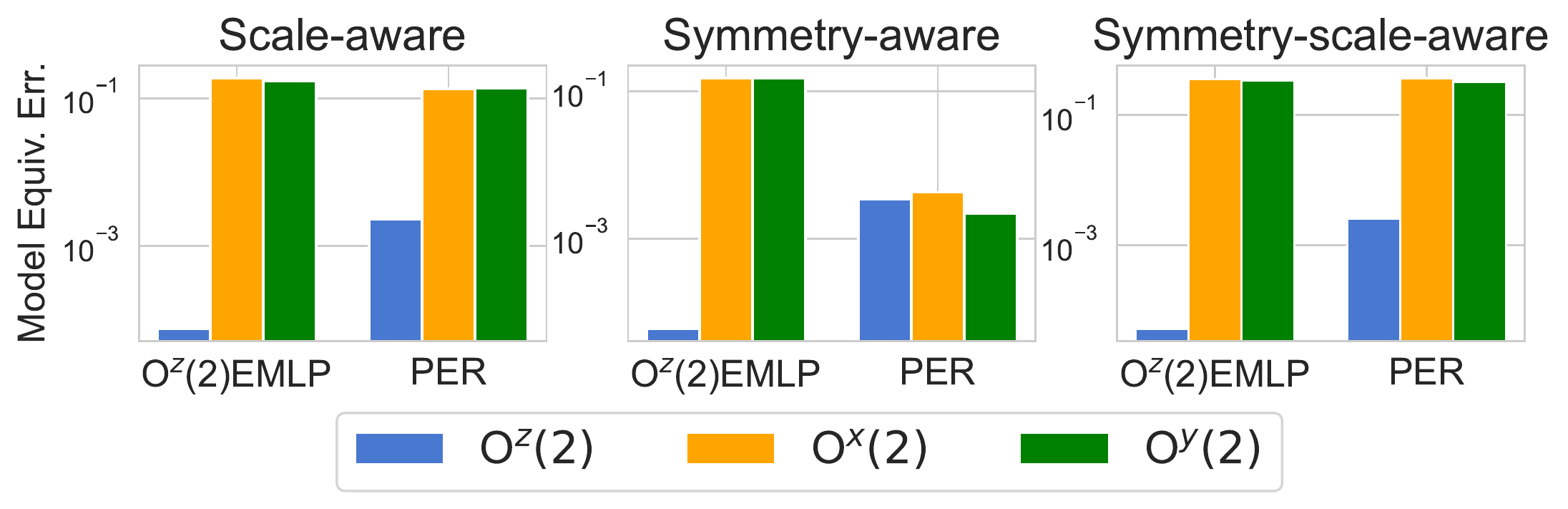}
    \caption{The model equivariance errors captured by $\gO^z(2)$-\gls{emlp} and our algorithm.}
    \label{fig:model-equiv}
 \end{figure}

We expect that the scale-aware normalization breaks the $\gO^z(2)$ equivariance because it normalizes the $x$ and $y$ axes with different scales, but the degree of approximate equivariance would not be serious because the $x$ axis and the $y$ axis have similar (but still different) scales. Indeed, \cref{fig:test-motion} shows that the models (approximately) equivariant to $\gO^z(2)$, $\gO^z(2)$-\gls{emlp}, $\gO^z(2)$-$\gO(3)$ \gls{memlp}, and \gls{per}, perform better than the others. Interestingly, as can be seen in \cref{fig:model-equiv}, \gls{per} discovers that the data has soft $\gO^z(2)$ equivariance, which coincides with our expectation that the scale-aware normalization mildly breaks the $\gO^z(2)$ equivariance. Note that $\gO^z(2)$-\gls{emlp} exhibits a tiny equivariance error. This is due to a numerical error in calculating the equivariant basis $Q$ and $R$ in~\cref{eq:definition-emlp}.

 Even though the symmetry-aware normalization does not break the $\gO(3)$ equivariance, the dataset itself has soft $\gO(3)$ equivariance due to the gravity acting on the vehicles. However, the significantly small scale of $z$-coordinates in the symmetry-aware normalization causes a model to underestimate the $\gO^x(2)$ and $\gO^y(2)$ equivariance. Consequently, the small equivariance error discovered by \gls{per} led to the best performance. As shown in~\cref{fig:model-equiv}, while $\gO^z(2)$-\gls{emlp} captures only large equivariance errors on $\gO^x(2)$ and $\gO^y(2)$, \gls{per} captures small equivariance errors on $\gO(3)$.
 
 For the symmetry-scale-aware scheme, \gls{per} shows the best performance. As in the scale-aware, all models perform well except for the $\gO(3)$-\gls{emlp}. Together with the captured equivariance errors in~\cref{fig:model-equiv}, they explain the symmetry-scale-aware scheme is also softly $\gO(3)$ equivariant. Whereas the element-wise scaling causes soft $\gO(3)$ equivariance in the scale-aware, in the symmetry-scale-aware scheme, mainly the gravity acting on the vehicles along the $z$-axis results in the soft $\gO(3)$ equivariance of the data. Moreover, the relatively large equivariance errors on $\gO^z(2)$ (blue) helped the performance of \gls{per}, which was a coherent result with~\citet{assaad2022vntransformer}. 

To summarize, for all three normalization, \gls{per} robustly outperforms the baselines, and discovers reasonable soft symmetries. 

%% file: tables/testmse_inertia.tex
\begin{table}
\small
 \setlength{\tabcolsep}{2pt}
\caption{Test MSE for the moment of inertia task. \gls{emlp} and \gls{rpp} are built with $\gO(3)$ and \gls{memlp} is built with $\gO(3)$-\gls{emlp} and $\gO^{(\text{ax})}$-\gls{emlp} where $\text{ax} \in \{x,y,z\}$.}
 \label{tab:test-inertia}
    \centering
 \resizebox{.45\textwidth}{!}{
    \begin{tabular}{ccccccc}\toprule
        Equiv group & MLP & $\gO(2)$\gls{emlp} &$\gO(3)$\gls{emlp} & \gls{rpp} & \gls{memlp} & \gls{per} \\
        \midrule
        $\gO(3)$ &$4.25\spm{0.17}$& - & $1.13\spm{0.36}$ & $2.66 \spm{1.43}$ & - & $\mathbf{0.27\spm{0.23}}$ \\
        $\gO^x(2)$ &$2.84\spm{0.12}$&$1.75\spm{0.63}$& $62.36 \spm 41.10$  & $2.06 \spm 1.12$  & $3.38 \spm 0.92$ & $\mathbf{0.25 \spm{0.16}}$ \\
        $\gO^y(2)$ &$2.78\spm{0.12}$&$1.73\spm{0.11}$&  $29.11 \spm 14.06$ & $1.72 \spm 0.61$ & $2.87 \spm 0.31$ &  $\mathbf{0.56\spm{0.48}}$\\
        $\gO^z(2)$ &$2.69\spm{0.10}$&$1.56\spm{0.17}$&  $46.32 \spm 10.18$ & $1.86 \spm 1.13$ & $2.75 \spm 0.29$ & $\mathbf{0.32\spm 0.26}$\\
        - &$6.81\spm{0.23}$& - & $10.65 \spm 2.08$ & $4.16 \spm 0.49$ & -  & $\mathbf{0.34\spm 0.28}$\\        
        \bottomrule
    \end{tabular}    
    }
\end{table}

%% file: tables/testmse_cossim.tex

\begin{table}
\small
 \setlength{\tabcolsep}{2pt}
\caption{Test MSE for the CosSim task. \gls{emlp} and \gls{rpp} are built with $(\gSO(3),\gS(3))$ and \gls{memlp} is built with $(\gSO(3),\gS(3))$-\gls{emlp} and $\gSO(3)$ or $\gS(3)$\gls{emlp} \gls{emlp}. Sub-\gls{emlp} stands for either $\gSO(3)$ or $\gS(3)$ and \gls{emlp} stands for $(\gSO(3),\gS(3))$-\gls{emlp}. All values are in a scale of $\times10^{-1}$.}
 \label{tab:test-cossim}
    \centering
    \resizebox{.45\textwidth}{!}{
    \begin{tabular}{ccccccc}\toprule
         Inv group & MLP & Sub-\gls{emlp} & \gls{emlp} & \gls{rpp} & \gls{memlp} & \gls{per} \\
        \midrule
        ${\gSO(3),\gS(3)}$ & $0.41\spm{0.03}$& - & $1.10\spm{0.02}$ & $1.10 \spm{0.03}$ & - & $\mathbf{0.32\spm{0.02}}$ \\
        $\gSO(3)$ &$0.46\spm{0.07}$&$\mathbf{0.39\spm{0.30}}$& $2.54 \spm 0.10$  & $2.57 \spm 0.10$  & $2.56 \spm 0.10$ & $0.44 \spm{0.03}$ \\
        $\gS(3)$ &$0.69\spm{0.04}$&$2.14\spm{0.11}$& $2.14 \spm 0.11$ & $2.18 \spm 0.09$ & $2.18 \spm 0.09$ &  $\mathbf{0.65\spm{0.09}}$\\ 
        - &$3.76\spm{0.32}$& - & $3.76 \spm 0.32$ & $3.84 \spm 0.04$ & - & $\mathbf{0.66\spm 0.13}$\\
        \bottomrule
    \end{tabular}    
    }
\end{table}

%% file: main/related_works.tex
\section{Related Work}
\label{main:sec:related_works}
The translation equivariance of CNN and permutation equivariance of GNN are the most popular examples of symmetry built in the neural networks. Recently, there have been several works designing neural networks having desirable group equivariance. \gls{emlp}~\citep{finzi2021practical} is a  framework that builds an MLP equivariant to various groups. LieConv~\citep{finzi2020generalizing} is a variant of CNN targeting equivariance for Lie groups. Another variant called $G$-CNN~\citep{cohen2016group} is equivariant w.r.t. 90-degree rotations, reflections, and translation.

Most softly equivariant models impose architectural restrictions for the soft equivariance. \gls{rpp}~\citep{finzi2021residual} build a soft equivariant model via a residual layer added to the equivariant linear layer, where the degree of equivariance is determined by the prior variances assigned for the equivariant layer and the residual pathway. Relaxed group convolution~\citep{wang2022approximately} implements a softly equivariant CNN by interpolating multiple conv operations with different weights, and the number of convolutions determines the degree of equivariance. Relaxed $G$-steerable group convolution~\citep{wang2022approximately} introduces spatial-location-dependent weights that replace the weights in the $G$-steerable CNN. Relaxed $G$- and $G$-steerable CNNs use group-action-based regularizers to restrict the relaxation.

There are some previous works allowing automatic symmetry discovery from data~\citep{dehmamy2021automatic,krippendorf2021detecting}.
However, to our knowledge, ours is the first to discover the varying degrees of
approximate equivariance across multiple groups under mixed symmetry settings.

%% file: main/conclusion.tex
\section{Conclusion}
\label{main:sec:conclusion}

In this paper, we tackle the learning problems under mixed symmetries, where a dataset contains multiple types of symmetries with different levels of equivariance errors. While previous methods focused on a single type of symmetries and bake in the equivariance constraint to the architecture as an inductive bias, ours take a regularizer-based approach, where a model without any equivariance constraint is regularized towards it using a projection-based regularization. One notable advantage is that it can automatically detect the levels of equivariance errors and adapt to those error levels by controlling the regularization coefficients. This is done during the training without any explicit supervision. Using a synthetic function approximation task and real-world motion forecasting task, we demonstrate that our proposed model could indeed capture mixed symmetries, identify the different level of equivariance errors, and predicts better than the existing methods. In this paper, we mainly focused on MLP architectures, so extending our framework to arbitrary neural network architectures such as CNNs, RNNs, or transformers~\citep{vaswani2017attention} would be an interesting future research direction.


%% file: appendix/proof.tex
\section{Proof of \mathwrap{\cref{prop:inequality}}}
\label{app:sec:proof}
\begin{proof}
We define notations $V$ and $\dag$ where $V$ is a vectorized form of the weight $W$ and $\dag$ converts the vector form of a matrix back to the matrix form. i.e. $V=\mathrm{vec}(W)$ and $W=V^\dag$. Also, we can utilize those identities by the definitions of $Q_w$ and $Q_b$.
\[
\rho_\calY(g)(Q_wQ_w^\intercal V)^\dag &= (Q_wQ_w^\intercal V)^\dag\rho_\calX(g), & \rho_\calY(g) Q_b Q_b^\intercal b &= Q_b Q_b^\intercal b.
\]

Now we first prove \cref{prop:inequality} where $f$ is a linear function. By the triangle inequality,
\[
\lefteqn{\norm{\rho_\calY(g)(W\bx+b)-W\rho_\calX(g)\bx-b}}\\
&=\lVert\rho_\calY(g)W\bx-\rho_\calY(g)(Q_wQ_w^\intercal V)^\dag\bx+\rho_\calY(g)b-\rho_\calY(g)Q_bQ_b^\intercal b\\
&\hspace{20pt}-W\rho_\calX(g)\bx+(Q_wQ_w^\intercal V)^\dag\rho_\calX(g)\bx-b+Q_bQ_b^\intercal b\rVert\\
&\leq\norm{\rho_\calY(g)W\bx-\rho_\calY(g)(Q_wQ_w^\intercal V)^\dag\bx+\rho_\calY(g)b-\rho_\calY(g)Q_bQ_b^\intercal b}\\
&\hspace{20pt}+\norm{W\rho_\calX(g)\bx-(Q_wQ_w^\intercal V)^\dag\rho_\calX(g)\bx+b-Q_bQ_b^\intercal b}\\
&\leq\norm{\rho_\calY(g)\big(W-(Q_wQ_w^\intercal V)^\dag\big)\bx}+\norm{\rho_\calY(g)(b-Q_bQ_b^\intercal b)}\\
&\hspace{20pt}+\norm{\big(W-(Q_wQ_w^\intercal V)^\dag\big)\rho_\calX(g)\bx}+\norm{b-Q_bQ_b^\intercal b}.
\]

We can split out $\rho(g)$ and $\bx$ by using the operator norms and the operator norm is bounded by Frobenius norm.
\[
\norm{\rho_\calY(g)\big(W-(Q_wQ_w^\intercal V)^\dag\big)\bx}&\leq\norm{\rho_\calY(g)}_\mathrm{op}\norm{W-(Q_wQ_w^\intercal V)^\dag}_F\norm{\bx},\\
\norm{\rho_\calY(g)(b-Q_bQ_b^\intercal b)}&\leq\norm{\rho_\calY(g)}_\mathrm{op}\norm{b-Q_bQ_b^\intercal b},\\
\norm{\big(W-(Q_wQ_w^\intercal V)^\dag\big)\rho_\calX(g)\bx}&\leq\norm{W-(Q_wQ_w^\intercal V)^\dag}_F\norm{\rho_\calX(g)\bx}.
\]
Therefore, the $G$-equivariance error is bounded as follows:
\[
\lefteqn{\sup_{\bx,g}\norm{\rho_\calY(g)(W\bx+b)-W\rho_\calX(g)\bx-b}}\\
&\leq\big(\sup_g\norm{\rho_\calY(g)}_\mathrm{op}\sup_{\bx}\norm{\bx}+\sup_{\bx,g}\sup_g\norm{\rho_\calX(g)\bx}\big)\norm{V-Q_wQ_w^\intercal V}\\
&\hspace{20pt}+\big(\norm{\rho_\calY(g)}_\mathrm{op}+1\big)\norm{b-Q_bQ_b^\intercal b}\\
\label{eq:bound-linear}&=C_1\norm{V-Q_wQ_w^\intercal V}+C_2\norm{b-Q_bQ_b^\intercal b}.
\]
The norm of $\bx$ are supposed to be bounded since we have a finite dataset. Now we are looking at when $f$ is a non-linear function whose activation $\sigma$ is $G$-equivariant and $L$-lipchitz continuous. The equivariant activation $\sigma$ has
\[
\rho_\calY(g)\sigma(f(\bx))=\sigma(\rho_\calY(g)f(\bx))
\] for any function $f$. Hence, the $G$-equivariance error
\[
\norm{\rho_\calY(g)\sigma(W\bx+b)-\sigma(W\rho_\calX(g)\bx-b)}=\norm{\sigma(\rho_\calY(g)(W\bx+b))-\sigma(W\rho_\calX(g)\bx+b)}.
\] Since $\sigma$ is $L$-lipchitz continuous,
\[
\norm{\sigma(\rho_\calY(g)(W\bx+b))-\sigma(W\rho_\calX(g)\bx+b)}\leq L\norm{\rho_\calY(g)(W\bx+b)-W\rho_\calX(g)\bx-b}.
\] The r.h.s is the equivariance error when $f$ is a linear function, which is bounded by \cref{eq:bound-linear}.

Lastly, we show the case when $f$ is a two-layer MLP. More-than-two-layered MLPs can be shown in the same way. The G-equivariance error is
\[
\lefteqn{\norm{\rho^{(2)}(g)(W^{(2)}\sigma(W^{(1)}\bx+b^{(1)})+b^{(2)})-W^{(2)}\sigma(W^{(1)}\rho^{(0)}(g)\bx+b^{(1)})-b^{(2)}}}\\
&=\nonumber\lVert\rho^{(2)}(g)(W^{(2)}\sigma(W^{(1)}\bx+b^{(1)})+b^{(2)})-W^{(2)}\rho^{(1)}(g)\sigma(W^{(1)}\bx+b^{(1)})-b^{(2)}\\
&\hspace{20pt}+W^{(2)}\rho^{(1)}(g)\sigma(W^{(1)}\bx+b^{(1)})+b^{(2)}-W^{(2)}\sigma(W^{(1)}\rho^{(0)}(g)\bx+b^{(1)})-b^{(2)}\rVert.
\]
This is bounded by an addition of two equivariance errors by triangle inequality
\[
\lefteqn{\norm{\rho^{(2)}(g)(W^{(2)}\sigma(W^{(1)}\bx+b^{(1)})+b^{(2)})-W^{(2)}\rho^{(1)}(g)\sigma(W^{(1)}\bx+b^{(1)})-b^{(2)}}}\nonumber\\
&\hspace{20pt}+\norm{W^{(2)}\rho^{(1)}(g)\sigma(W^{(1)}\bx+b^{(1)})+b^{(2)}-W^{(2)}\sigma(W^{(1)}\rho^{(0)}(g)\bx+b^{(1)})-b^{(2)}}\\
&\leq\nonumber\norm{\rho^{(2)}(g)(W^{(2)}\bx'+b^{(2)})-W^{(2)}\rho^{(1)}(g)\bx'-b^{(2)}}\\
&\label{eq:bound-mlp1}\hspace{20pt}+\norm{W^{(2)}}_\mathrm{op}\norm{\rho^{(1)}(g)\sigma(W^{(1)}\bx+b^{(1)})-\sigma(W^{(1)}\rho^{(0)}(g)\bx+b^{(1)})},
\] where $\bx'=\sigma(W^{(1)}\bx+b^{(1)})$. The first term of \cref{eq:bound-mlp1} is the equivariance error of the linear function where the input is the output of the first layer. Besides, the second term involves the equivariance error of the non-linear function. Overall, the equivariance error of the two-layer MLP is bounded as
\[
\lefteqn{\sup_{\bx,g}\norm{\rho^{(2)}(g)(W^{(2)}\sigma(W^{(1)}\bx+b^{(1)})+b^{(2)})-W^{(2)}\sigma(W^{(1)}\rho^{(0)}(g)\bx+b^{(1)})-b^{(2)}}}\\
&\leq \big(\sup_g\norm{\rho^{(2)}(g)}_\mathrm{op}\sup_\bx\norm{\bx'}+\sup_{\bx,g}\norm{\rho^{(1)}(g)\bx'}\big)\norm{V^{(2)}-Q_{w^{(2)}}Q_{w^{(2)}}^\intercal V^{(2)}}\\
&\hspace{20pt}+\big(\sup_g\norm{\rho^{(2)}(g)}_\mathrm{op}+1\big)\norm{b^{(2)}-Q_{b^{(2)}}Q_{b^{(2)}}^\intercal b^{(2)}}\\
&\hspace{20pt}+L\norm{W^{(2)}}_\mathrm{op}\big(\sup_g\norm{\rho^{(1)}(g)}_\mathrm{op}\sup_\bx\norm{\bx}+\sup_{\bx,g}\norm{\rho^{(0)}(g)\bx}\big)\norm{V^{(1)}-Q_{w^{(1)}}Q_{w^{(1)}}^\intercal V^{(1)}}\\
&\hspace{20pt}+L\norm{W^{(2)}}_\mathrm{op}\big(\sup_g\norm{\rho^{(1)}(g)}_\mathrm{op}+1\big)\norm{b^{(1)}-Q_{b^{(1)}}Q_{b^{(1)}}^\intercal b^{(1)}}\\
&\leq C_1^{(2)}\norm{V^{(2)}-Q_{w^{(2)}}Q_{w^{(2)}}^\intercal V^{(2)}}+C_2^{(2)}\norm{b^{(2)}-Q_{b^{(2)}}Q_{b^{(2)}}^\intercal b^{(2)}}\\
&\hspace{20pt}+C_1^{(1)}\norm{V^{(1)}-Q_{w^{(1)}}Q_{w^{(1)}}^\intercal V^{(1)}}+C_2^{(1)}\norm{b^{(1)}-Q_{b^{(1)}}Q_{b^{(1)}}^\intercal b^{(1)}}.
\]
In terms of the mathematical induction, the bound of the equivariance error of more-than-two-layered MLPs can be derived as follows:
\[
\lefteqn{\sup_{\bx,g}[\norm{\rho^{(S)}(g)(W^{(S)}\sigma(f'(\bx))+b^{(S)})-W^{(S)}\sigma(f'(\rho^{(0)}(g)\bx))-b^{(S)}}]}\\
&=\nonumber\sup_{\bx,g}\lVert\rho^{(S)}(g)(W^{(S)}\sigma(f'(\bx))+b^{(S)})-W^{(S)}\rho^{(1)}(g)\sigma(f'(\bx))-b^{(S)}\\
&\hspace{20pt}+W^{(S)}\rho^{(S-1)}(g)\sigma(f'(\bx))+b^{(S)}-W^{(S)}\sigma(f'(\rho^{(0)}(S)\bx))-b^{(S)}\rVert\\
&\leq\nonumber\sup_{\bx,g}\lVert\rho^{(S)}(g)(W^{(S)}\sigma(f'(\bx))+b^{(S)})-W^{(S)}\rho^{(1)}(g)\sigma(f'(\bx))-b^{(S)}\rVert\\
&\hspace{20pt}+L\norm{W^{(S)}}_\mathrm{op}\sup_{\bx,g}\lVert\rho^{(S-1)}(g)f'(\bx)-f'(\rho^{(0)}(S)\bx)\rVert,
\] where $S$ is the number of layers and $f'$ is a $(S-1)$-layered MLP.
\end{proof}

%% file: appendix/rppvsper.tex
\section{RPP vs. PER}
\label{app:sec:compare}
\begin{figure}[h]
    \label{fig:parameter}
    \centering
    \includegraphics[width = 0.27\textwidth]{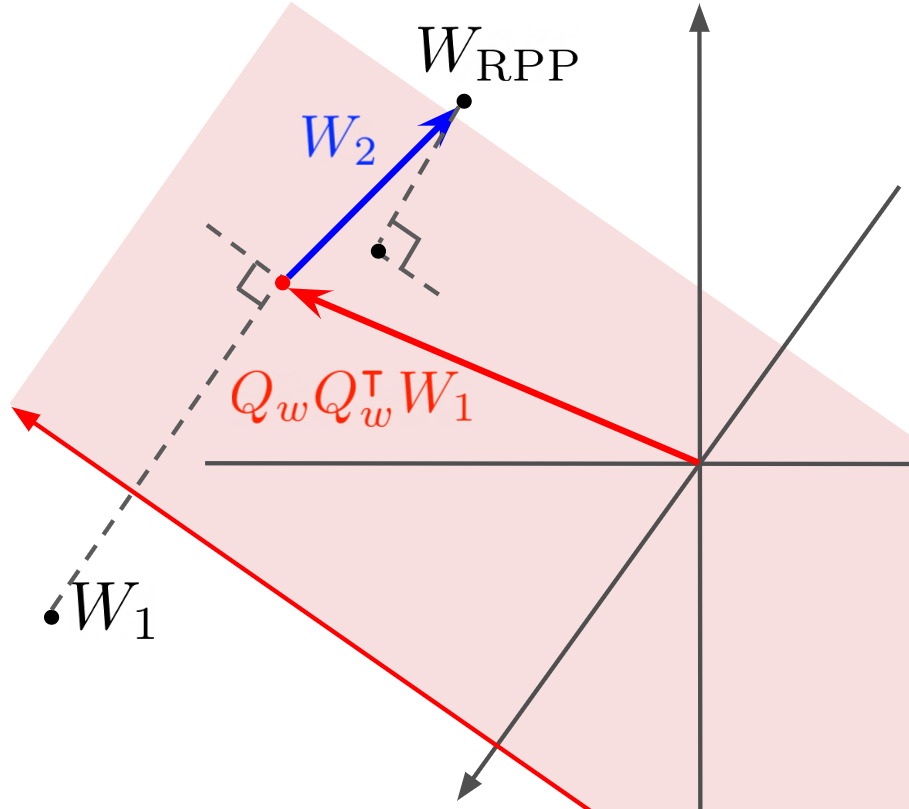}
    \includegraphics[width = 0.27\textwidth]{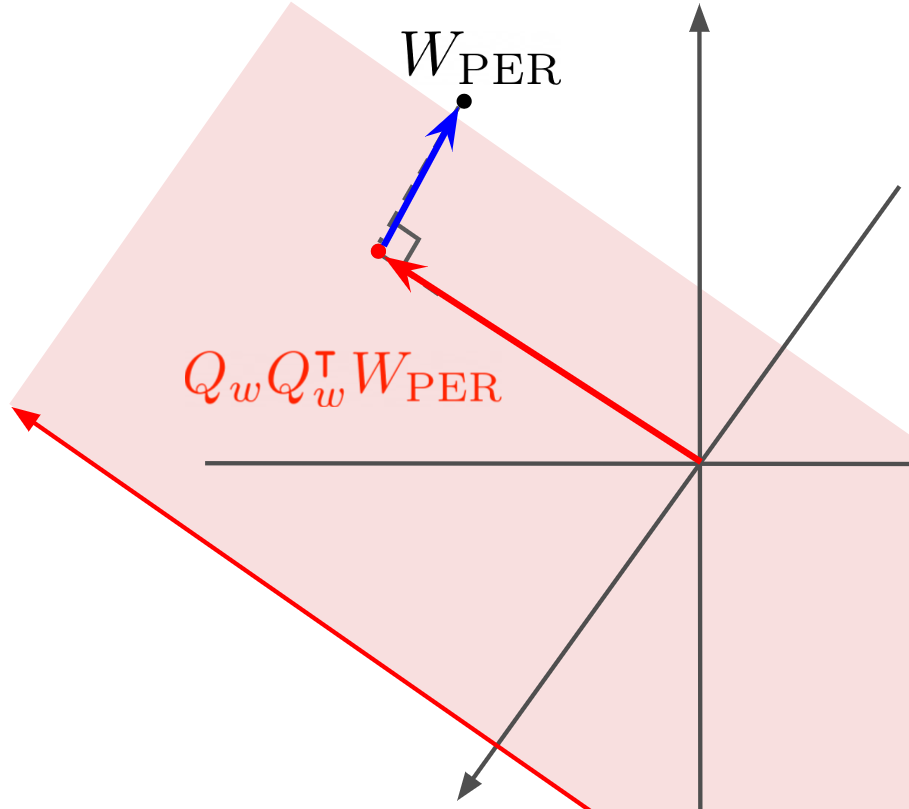}
    \caption{Comparison of parameterization bewteen \gls{rpp} (left) and \gls{per} (right). $W_1$ and $W_2$ are the parameters of \gls{rpp}. $W_1$ explains equivariance by projecting onto the equivariant space and $W_2$, called a residual path, captures the difference between approximate equivariance desired in dataset and strict equivariance of $Q_wQ_w^\intercal W_1$. On the other hand, $W_\text{PER}$ does not require additional parameters because it is already close to the equivariant space due to the regularizer.}
    \label{fig:rpp vs per}
 \end{figure}

Illustrated in~\cref{fig:rpp vs per}.


%% file: appendix/initializations.tex
\section{Weight Initializations for PER}
\label{app:sec:initialization}
Since our method does not restrict the parameter space, we can freely choose desirable strategies of weight initialization according to prior knowledge about a given task.
\subsection{Standard}
Obviously we can utilize the well-known initializations of neural networks such as Glorot initialization~\citep{glorot2010understanding} and He initialization~\citep{he2015delving}.

\subsection{Soft}
This initialization mimics the initial weights of the RPP model. The structure of the RPP models consists of addition of weights projected on the equivariant space $QQ^\intercal\text{vec}(W_1)$ and small weights $\text{vec}(W_2)$ acting as a perturbation of the equivariant weights.
\[
\text{vec}(W_\text{RPP}) &= QQ^\intercal\text{vec}(W_1)+\text{vec}(W_2), & \text{vec}(W_1)&\sim\normal(0,\sigma^2\bI), & \text{vec}(W_2)&\sim\normal(0,\epsilon\sigma^2\bI),
\] where $0<\epsilon<<1$ and $\sigma$ is determined by selected types of initialization such as Glorot and He.
Thus, our model can be initialized with the added distribution as follows:
\[
\label{eq:soft-init}
\text{vec}(W_\text{PER})\sim\normal(0, \sigma^2QQ^\intercal+\epsilon\sigma^2\bI).
\]

\subsection{Half Soft}
The degree of approximate equivariance is determined by the perpendicular distance from the equivariant space and the perpendicular distance is determined by an amount of the complementary direction of the equivariant space. i.e. the approximate equivariance degree of weight $W$ is determined by $\tilde{Q}\tilde{Q}^\intercal\text{vec}(W)$ because $\text{vec}(W) = QQ^\intercal \text{vec}(W)+\tilde{Q}\tilde{Q}^\intercal \text{vec}(W)$, where $\tilde{Q}$ is the complementary basis of $Q$. Therefore, we can control the equivariance of the initial weights with a scaling factor $\lambda$ as follows:
\[
\text{vec}(W_\text{PER})\sim\normal(0, (1-\lambda)\sigma^2QQ^\intercal+\lambda\sigma^2\bI)=\normal(0, \sigma^2QQ^\intercal+\lambda\sigma^2\tilde{Q}\tilde{Q}^\intercal).
\]
The case when $\lambda=0$ corresponds to the initial weights of EMLP and the case when $\lambda=1$ corresponds to the initial weights of MLP. We chose $\lambda=0.5$ to locate the model in the middle between EMLP and MLP.

%% file: appendix/correlation.tex
\section{Samples for Measuring Correlation}
\label{app:sec:correlation}
Experiments for measuring the Pearson correlation are listed in~\cref{tab:correlation}.

\input{tables/correlation.tex}

%% file: tables/correlation.tex
\begin{table}[ht]
    \caption{Samples for measuring the correlation with the data equivariance error. $\epsilon_1=-\calI\hat{x}\hat{x}^\intercal$, $\epsilon_2=-\calI\hat{y}\hat{y}^\intercal$, $\epsilon_3=-\calI\hat{z}\hat{z}^\intercal$, and $\epsilon_4=-\calI\hat{x}\hat{x}^\intercal +\calI\hat{y}\hat{y}^\intercal-\calI\hat{z}\hat{z}^\intercal$}
    \centering
    \vskip 0.05in
    \scriptsize
    \def\arraystretch{1.1}
        \begin{tabular}{lrrrrrrrrrrrr}
            \toprule
            & \multicolumn{3}{c}{Data Equiv. Err.} & \multicolumn{3}{c}{Equiv. Regular. Coeff.} & \multicolumn{3}{c}{Model Equiv. Err.} & \multicolumn{3}{c}{Equiv. Regular.} \\
            \cmidrule(lr){2-4}\cmidrule(lr){5-7}\cmidrule(lr){8-10}\cmidrule(lr){11-13}
            Noise & $\text{O}^z(2)$ & $\text{O}^x(2)$ & $\text{O}^y(2)$ & $\text{O}^z(2)$ & $\text{O}^x(2)$ & $\text{O}^y(2)$ & $\text{O}^z(2)$ & $\text{O}^x(2)$ & $\text{O}^y(2)$ & $\text{O}^z(2)$ & $\text{O}^x(2)$ & $\text{O}^y(2)$ \\
            \midrule
            0 & 7.22E-08 & 7.23E-08 & 7.38E-08 & 1.00E+02 & 9.80E+01 & 9.34E+01 & 9.17E-04 & 7.72E-04 & 7.60E-04 & 1.46E-07 & 1.32E-07 & 1.31E-07 \\
            $0.3\epsilon_1$ & 5.89E-02 & 7.05E-08 & 5.85E-02 & 1.35E+01 & 1.00E+02 & 1.32E+01 & 5.30E-02 & 4.27E-03 & 5.42E-02 & 7.06E-05 & 9.95E-07 & 7.01E-05 \\
            $0.6\epsilon_1$ & 1.05E-01 & 7.11E-08 & 1.04E-01 & 1.22E+01 & 1.00E+02 & 1.29E+01 & 7.56E-02 & 8.09E-03 & 7.95E-02 & 1.19E-04 & 4.78E-06 & 1.18E-04 \\
            $0.9\epsilon_1$ & 1.41E-01 & 6.90E-08 & 1.40E-01 & 1.97E+01 & 1.00E+02 & 1.98E+01 & 1.36E-01 & 3.78E-03 & 1.34E-01 & 9.61E-05 & 2.20E-06 & 9.53E-05 \\
            $0.3\epsilon_2$ & 5.99E-02 & 5.34E-02 & 7.43E-08 & 7.54E+00 & 7.36E+00 & 1.00E+02 & 4.54E-02 & 5.06E-02 & 9.64E-03 & 1.85E-04 & 1.84E-04 & 1.57E-05 \\ 
            $0.6\epsilon_2$ & 1.08E-01 & 9.74E-02 & 7.22E-08 & 2.64E+01 & 2.62E+01 & 1.00E+02 & 8.13E-02 & 8.71E-02 & 1.73E-02 & 1.28E-04 & 1.27E-04 & 2.62E-05 \\
            $0.9\epsilon_2$ & 1.47E-01 & 1.34E-01 & 7.15E-08 & 9.61E+00 & 9.86E+00 & 1.00E+02 & 1.08E-01 & 1.10E-01 & 7.21E-03 & 2.44E-04 & 2.44E-04 & 1.96E-05 \\
            $0.3\epsilon_3$ & 7.04E-08 & 5.34E-02 & 5.96E-02 & 1.00E+02 & 3.36E+00 & 3.39E+00 & 1.67E-03 & 5.55E-02 & 5.56E-02 & 7.64E-07 & 1.41E-04 & 1.41E-04 \\
            $0.6\epsilon_3$ & 7.05E-08 & 9.75E-02 & 1.08E-01 & 1.00E+02 & 4.74E+00 & 4.96E+00 & 2.02E-03 & 9.61E-02 & 9.47E-02 & 2.87E-06 & 2.04E-04 & 2.05E-04 \\
            $0.9\epsilon_3$ & 6.88E-08 & 1.34E-01 & 1.47E-01 & 1.00E+02 & 4.19E+00 & 4.21E+00 & 1.87E-03 & 1.39E-01 & 1.40E-01 & 2.40E-06 & 2.65E-04 & 2.65E-04\\
            $0.3\epsilon_4$ & 1.32E-01 & 5.91E-02 & 6.70E-02 & 1.79E+01 & 9.73E+01 & 1.00E+02 & 1.17E-01 & 5.13E-02 & 6.59E-02 & 9.04E-05 & 3.46E-05 & 3.39E-05 \\
            $0.6\epsilon_4$ & 2.50E-01 & 1.14E-01 & 1.31E-01 & 3.68E+01 & 9.62E+01 & 1.00E+02 & 1.96E-01 & 9.09E-02 & 1.06E-01 & 1.53E-04 & 5.79E-05 & 5.92E-05 \\
            $0.9\epsilon_4$ & 3.41E-01 & 1.58E-01 & 1.85E-01 & 3.29E+01 & 5.17E+01 & 1.00E+02 & 3.07E-01 & 1.39E-01 & 1.75E-01 & 1.86E-04 & 6.21E-05 & 5.86E-05 \\
            \bottomrule
        \end{tabular}
    \label{tab:correlation}
\end{table}

%% file: appendix/trajectory_comparison.tex
\section{Comparison of trajectory between the normalizations}
\label{app:sec:trajectory-comparison}
\begin{figure}[h]
     \centering
     \includegraphics[width=\textwidth]{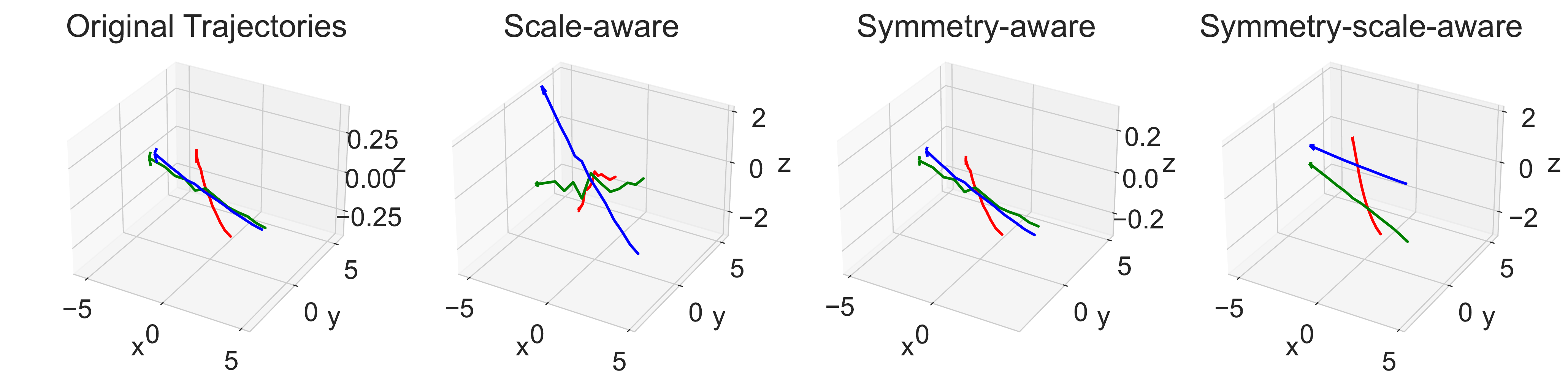}
     \caption{ The scale-aware normalization strongly emphasizes the $z$-coordinates. The symmetry-aware normalization just scales down the whole coordinates but the scale of $z$-coordinates is still close to zero. The symmetry-scale-aware normalization scales down the whole coordinates while retaining the scale of $z$-coordinates.}
     \label{fig:trajectory}
 \end{figure}

3 example trajectories (red, green, and blue) for each normalization are described in~\cref{fig:trajectory}.

%% file: appendix/details.tex
\section{Experimental Details}
\label{app:sec:details}

\subsection{Dataset Description}
Information for each dataset is summarized in~\cref{tab:dataset_statistics}.
\input{tables/dataset_statistics}

\subsection{Data Selection of WOMD}
\paragraph{Trajectory Slicing} The WOMD dataset contains maximum 91 points of a trajectory measured in 10Hz. We sliced and gathered first 24 points and dropped every even-numbered points so that the final trajectory contains only 12 points. The past 6 points and future 6 points are regarded as input and output, respectively.

\paragraph{Trajectory Selection}
We selected only a portion of the whole WOMD dataset. The training part of WOMD motion forecasting dataset consists of
total 1,000 files of the \texttt{TFRecord} format. We used first 28 files as training set, next 6 files as validation set, and last 6 files as testing set only. Furthermore, we excluded all trajectory that doesn't move enough and move too far, we collected only trajectories satisfying the following conditions:
\[
\norm{\by^{t=6}\cdot\hat{x}-\by^{t=1}\cdot\hat{x}}_2 &< 5\\
\norm{\by^{t=6}\cdot\hat{y}-\by^{t=1}\cdot\hat{y}}_2 &< 5\\
\norm{\by^{t=6}\cdot\hat{z}-\by^{t=1}\cdot\hat{z}}_2 &> 0.05,
\] where $\hat{x},\hat{y},\hat{z}\in\bbR^3$ are orthonormal basis vectors of $x$-,$y$-, and $z$-axes.

\subsection{Details of Training}
For all experiments, we used five different seeds to report performance results.
\paragraph{Architecture Description}
All architectures of the neural networks including EMLP, RPP, Mixed RPP, and our model are fixed with 4 layers and different width. Their widths were adjusted to set their number of parameters the same.

\paragraph{Hyperparameters}
See \cref{tab:commom-hyperparam} for hyperparameter settings of our model and baseline methods for all experiments. Those hyperparameters are applied the same for all models. Additional hyperparameters of our model for each task are listed in~\cref{tab:per-hyperparam}.
\input{tables/common_hyperparam.tex}
\input{tables/per_hyperparam.tex}

\paragraph{Extra Details}
We applied the cosine decaying of learning rate~\citep{loshchilov2017sgdr} and early stopping with 50 patience for stable training. The optimizers used in every tasks are ADAM~\citep{kingma2015adam}. All experiments were trained and evaluated on RTX 3090 devices.

%% file: tables/dataset_statistics.tex
\begin{table}[ht]
    \caption{Information of each dataset. $\bS$ denotes a scalar and $\bV$ denotes a vector in $\bbR^3$.}
    \centering
    \vskip 0.05in
        \begin{tabular}{lrrr}
        \toprule
                                     & Inertia & CosSim & WOMD \\
        \midrule
        Training Samples            &        1,000 &    1,000 &           16,814 \\
        Validation Samples          &            1,000 &   1,000 &             3,339 \\
        Testing Samples &           1,000 &       1,000 &  3,563 \\
        Input Representation &    $5\bS\oplus5\bV$ &  $3\bV$ & $6\bV$ \\
        Output Representation &   $\bV^2$ &  $\bS$ & $6\bV$ \\
        \bottomrule
        \end{tabular}
    \label{tab:dataset_statistics}
\end{table}

%% file: tables/common_hyperparam.tex
\begin{table}[ht]
    \caption{Common hyperparameter settings for each task.}
    \centering
    \vskip 0.05in
    \def\arraystretch{1.1}
        \begin{tabular}{lrrrrr}
            \toprule
            Dataset & Mini-batch & Max Epochs & Learning Rate & Weight Decay & Width (RPP)\\
            \midrule
            \multirow{1}{*}{Inertia} & 500 & 8,000 & 0.001 & 2.0$\times10^{-4}$ & 384 (270)\\
            \multirow{1}{*}{CosSim}  & 200 &10,000 & 0.0002 & 2.0$\times10^{-5}$ & 128 (45)\\
            \multirow{1}{*}{WOMD (scale-aware)}    & 256 &   750 & 0.0002 & 0 & 384 (269) \\
            \multirow{1}{*}{WOMD (symmetry-aware)}    & 256 &   500 & 0.0002 & 0 & 384 (269) \\
            \multirow{1}{*}{WOMD (symmetry-scale-aware)}    & 256 &   500 & 0.0002 & 0 & 384 (269) \\
            \bottomrule
        \end{tabular}
    \label{tab:commom-hyperparam}
\end{table}

%% file: tables/per_hyperparam.tex
\begin{table}[h]
    \caption{Hyperparameter setting of our model.}
    \centering
    \vskip 0.05in
    \def\arraystretch{1.1}
        \begin{tabular}{llcccccc}
            \toprule
            Dataset & Task Type & Initial $\lambda$ & $\gamma$ & Adjustment Epoch & Initialization & Mini-batch & Max Epochs\\
            \midrule
            \multirow{5}{*}{Inertia}
            & $\text{O}(3)$ & 100 & 2 & 2,000 & Standard & 500 & 8,000\\
            & $\text{O}^x(2)$ & 100 & 2 & 2,000 & Standard & 500 & 8,000\\
            & $\text{O}^y(2)$ & 100 & 2 & 2,000 & Standard & 500 & 8,000\\
            & $\text{O}^z(2)$ & 100 & 2 & 2,000 & Standard & 500 & 8,000\\
            & Only Soft & 100 & 2 & 2,000 & Standard & 500 & 8,000\\
            \midrule
            \multirow{4}{*}{CosSim}
            & $\text{SO}(3) \cup \text{S}(3)$ & 0.005 & 2 & 2,500 & Standard & 200 & 10,000\\
            & $\text{SO}(3)$ & 0.1 & 2 & 2,500& Standard & 200 & 10,000\\
            & $\text{S}(3)$ & 0.01 & 2 & 2,500 & Standard & 200 & 10,000\\
            & Only Soft & 0.005 & 2 & 2,500 & Standard & 200 & 10,000\\
            \midrule
            \multirow{3}{*}{WOMD}
            & Scale-aware & 0.2 & 5 & 125 & Half Soft & 128 & 500\\
            & Symmetry-aware & 0.3 & 5 & 100 & Half Soft & 128 & 500\\
            & Symmetry-scale-aware & 5 & 5 & 100 & Half Soft & 128 & 500\\
            \bottomrule
        \end{tabular}
    \label{tab:per-hyperparam}
\end{table}

%% file: appendix/additional.tex
\section{Additional Experiments}
\label{app:sec:additional}

\subsection{Analysis of Adjustment of Hyperparameters}
\label{app:sec:hyperparameters}

We share a part of the robustness analysis across different hyperparameters (initial coefficients $\lambda$, scaling factors $\gamma$, and the moments of the adjustment) required in the automatic tuning procedure described in~\cref{main:sec:autotune}. As the results show in~\cref{tab:adjust}, we found that the performance of the model is not so sensitive to the choice of hyperparameters. For instance, for the initial value of lambda, we observed that the model would achieve similar performances provided that the ratio of the initial \emph{Loss} over the initial $\lambda\cdot\calR^\text{PER}$ is at a certain level. The situation for the scaling factor $\gamma$ is similar: the final performance was consistent for the values arbitrarily chosen within the range [2, 5].
\input{tables/adjustment_hyperparameters}

\subsection{Comparison with Frame Averaging}
\label{app:sec:fa}
\input{tables/fa} Frame Averaging (FA) \citep{puny2022frame} is a framework that, in simple terms, trains a 
$G$-equivariant model $f$ by taking the average over some group elements in $G$, called frames. FA is a flexible approach since it does not restrict the internal structure of the model $f$, unlike EMLP.

We ran additional experiments with FA on the fully-equivariant task same as the first row in~\cref{tab:test-inertia}. \cref{tab:fa} shows the results. Although EMLP in the table used gated nonlinearity (GNL) due to its architectural restriction, FA does not need such a restriction, so the "MLP w/ FA" row in~\cref{tab:fa} applied the frame averaging to the same setup as the MLP row (i.e., MLP with the Swish activation).

Our results confirmed, FA is indeed a more powerful baseline than EMLP. But note that our model (\gls{per}) performs better than MLP w/ FA here.

\subsection{Simple Experiment Assuming Symmetries Are Unknown in the WOMD task}
\input{tables/unknown} We explain an additional experiment where we mimic the situation of unknown symmetries by including various and sometimes wrong matrix groups as candidate groups and checking whether our method picks the correct groups. \cref{tab:unknown} shows the model equivariance error captured by the model when using all O(2), SL(2), and GL(2) \glspl{per} to train the motion forecasting task with symmetry-aware normalization (this task has symmetries with respect to O$^z$(2) and O$^x$(2)). As shown in the tables, our method has appropriately captured the equivariance with respect to O$^z$(2) and O$^x$(2).

%% file: tables/adjustment_hyperparameters.tex
\begin{table}[h]
\caption{Test MSE results across different hyperparameters required in the automatic \gls{per}-coefficients-tuning precedure described in~\cref{main:sec:autotune}. $\lambda$ is the initial coefficients and $\gamma$ is the scaling factors.}
\begin{subtable}[t]{0.16\textwidth}
\caption{Inertia O$^z$(2) task}
 \label{tab:adjust1}
    \centering
    \resizebox{\textwidth}{!}{
    \begin{tabular}{cc}\toprule
loss/($\lambda\cdot\calR^\text{PER}$) & Test MSE \\ \midrule
0.00009 & 1.75±2.27 \\
0.00037 & 0.32±0.26 \\
0.00147 & 0.35±0.19 \\
O$^z$(2)EMLP & 1.56±0.17    \\  
        \bottomrule
    \end{tabular}
    }
    \end{subtable}
\hfill
\begin{subtable}[t]{0.16\textwidth}
\caption{CosSim S(3) task}
 \label{tab:adjust2}
    \centering
    \resizebox{\textwidth}{!}{
    \begin{tabular}{cc}\toprule
Loss/($\lambda\cdot\calR^\text{PER}$) & Test MSE \\
\midrule
0.0348 & 0.068±0.007 \\
0.1741 & 0.065±0.009 \\
0.8707 & 0.052±0.013 \\
S(3)EMLP & 0.21±0.11   \\  
        \bottomrule
    \end{tabular}   
    }
    \end{subtable}
\hfill
\begin{subtable}[t]{0.16\textwidth}
\caption{Inertia O$^z$(2) task}
 \label{tab:adjust3}
    \centering
    \resizebox{\textwidth}{!}{
    \begin{tabular}{cc}\toprule
$\gamma$ & Test MSE \\
\midrule
2 & 0.32±0.26 \\
3 & 0.40±0.24 \\
4 & 0.35±0.15 \\
5 & 0.43±0.21 \\
O$^z$(2)EMLP & 1.56±0.17\\  
        \bottomrule
    \end{tabular}   
    }
    \end{subtable}
\hfill
\begin{subtable}[t]{0.16\textwidth}
\caption{CosSim S(3) task}
 \label{tab:adjust4}
    \centering
        \resizebox{\textwidth}{!}{
    \begin{tabular}{cc}\toprule
$\gamma$ & Test MSE \\
\midrule
2 & 0.065±0.009 \\
3 & 0.044±0.004 \\
4 & 0.044±0.004 \\
5 & 0.044±0.004 \\
S(3)EMLP & 0.214±0.110\\       
        \bottomrule
    \end{tabular}   
    }
    \end{subtable}
\hfill
\begin{subtable}[t]{0.16\textwidth}
\caption{Inertia O$^z$(2) task (training epochs 8000)}
 \label{tab:adjust5}
    \centering
        \resizebox{\textwidth}{!}{
    \begin{tabular}{cc}\toprule
  Adjusted Epoch & Test MSE \\
  \midrule
1000 & 0.26±0.17 \\
2000 & 0.32±0.26 \\
3000 & 0.38±0.24 \\
O$^z$(2) EMLP & 1.56±0.17\\       
        \bottomrule
    \end{tabular}
    }
    \end{subtable}
\hfill
\begin{subtable}[t]{0.16\textwidth}
\caption{CosSim S(3) task (training epochs 2000)}
\label{tab:adjust6}
    \centering
    \resizebox{\textwidth}{!}{
    \begin{tabular}{ccccc}\toprule
Adjusted Epoch & Test MSE \\
\midrule
300 & 0.044±0.003 \\
500 & 0.065±0.009 \\
700 & 0.046±0.004 \\
S(3) EMLP & 0.214±0.110 \\       
        \bottomrule
    \end{tabular}
}
    \end{subtable}
\label{tab:adjust}
\end{table}

%% file: tables/fa.tex
\begin{wraptable}{r}{4.5cm}
\small
\caption{Test MSE comparison with FA in the Inertia O(3) task}
    \centering
 \resizebox{.2\textwidth}{!}{
    \begin{tabular}{cc}\toprule
Model & Test MSE \\
\midrule
MLP & 4.25±0.17 \\
EMLP & 1.13±0.36 \\
RPP & 2.66±1.43 \\
PER & \textbf{0.27±0.23} \\
MLP w/ FA & 0.36±0.05\\        
        \bottomrule
    \end{tabular}    
    }
    \label{tab:fa}
\end{wraptable}

%% file: tables/unknown.tex
\begin{wraptable}{r}{0.6\textwidth}
\caption{(a) Captured equivariance errors across the prescribed regularizers with different groups (b) Change of Test MSE due to the additional regularizers (SL$^z$(2), SL$^y$(2), GL$^x$(2), and GL$^y$(2)).}
\begin{subtable}[t]{0.3\textwidth}
\caption{Model equivariance errors}
 \label{tab:unknown-error}
    \centering
    \resizebox{\textwidth}{!}{
    \begin{tabular}{cc}\toprule
Regularized Groups & Model Equiv. Err. \\
\midrule
O$^z$(2) & 0.0007 \\
O$^x$(2) & 0.0006 \\
SL$^z$(2) & 0.2638 \\
SL$^y$(2) & 0.2302 \\
GL$^x$(2) & 0.2398 \\
GL$^y$(2) & 0.2089    \\  
        \bottomrule
    \end{tabular}
    }
    \end{subtable}
\hfill
\begin{subtable}[t]{0.3\textwidth}
\caption{Test MSE}
 \label{tab:unknown-mse}
    \centering
    \resizebox{\textwidth}{!}{
    \begin{tabular}{cc}\toprule
Models & Test MSE (×10$^{-2}$) \\
\midrule
O(2) PER & 3.07±0.01 \\
O(2),SL(2),GL(2) PER & 3.09±0.01\\
        \bottomrule
    \end{tabular}   
    }
    \end{subtable}
\label{tab:unknown}
\end{wraptable}